\documentclass[runningheads]{llncs}
\usepackage{xcolor}
\usepackage[T1]{fontenc}
\usepackage{graphicx}
\usepackage{amsmath}
\usepackage{amsfonts}
\usepackage{enumitem}
\usepackage{setspace}
\usepackage{booktabs}
\usepackage{multirow}
\usepackage{adjustbox}
\usepackage{indentfirst}
\usepackage{enumitem}
\usepackage{hyperref}
\usepackage{microtype}
\usepackage{float}
\usepackage[ruled,vlined,linesnumbered]{algorithm2e}
\newcommand{\MP}[1]{\textcolor{red}{#1}}

\definecolor{besttime}{RGB}{0, 0, 255}    
\definecolor{bestmetric}{RGB}{255, 127, 0} 

\begin{document}
\title{SHARC: Reference point driven Spherical Harmonic Representation for Complex Shapes}
\titlerunning{SHARC}
%
\author{Panagiotis Sapoutzoglou \and
George Terzakis\and
Maria Pateraki}
\authorrunning{P. Sapoutzoglou et al.}
%
\institute{National Technical University of Athens, Athens, Greece \\
\email{\{psapoutzoglou,gterzakis,mpateraki\}@mail.ntua.gr}} 
%
\maketitle              
\begin{abstract}

We propose SHARC, a novel framework that synthesizes arbitrary, genus-agnostic shapes by means of a collection of Spherical Harmonic (SH) representations of distance fields. These distance fields are anchored at optimally placed reference points in the interior volume of the surface in a way that maximizes learning of the finer details of the surface. 
 To achieve this, we employ a cost function 
 that jointly maximizes sparsity and centrality in terms of positioning, as well as visibility of the surface from their location. For each selected reference point, we sample the visible distance field to the surface geometry via ray-casting and compute the SH coefficients using the Fast Spherical Harmonic Transform (FSHT). To enhance geometric fidelity, we apply a configurable low-pass filter to the coefficients and refine the output using a local consistency constraint based on proximity. Evaluation of SHARC against state-of-the-art methods demonstrates that the proposed method outperforms existing approaches in both reconstruction accuracy and time efficiency without sacrificing model parsimony. The source code is available at \url{https://github.com/POSE-Lab/SHARC}.


\end{abstract}

\section{Introduction}
Representing complex 3D shapes in a way that is both compact and capable of preserving fine geometric details remains a central challenge in computer graphics, geometry processing, and 3D generative modeling. Explicit representations, such as triangle meshes and point clouds, are widely adopted but 
their complexity often scales with detail, making them expensive to transmit, edit, or utilize in optimization tasks \cite{park2019_deepsdf,mescheder2019_occupancy}(e.g. neural rendering). On the other hand, implicit representations 
typically involve continuous functions and therefore offer the flexibility of reproducing representations such as point clouds or meshes by means of resampling.  
Modern approaches, such as neural fields, parameterize these functions using deep networks, often conditioned on low-dimensional latent codes, to achieve high-quality reconstruction and generative modeling \cite{park2019_deepsdf,mescheder2019_occupancy,Chen2019}. 
Despite their advantages, these methods encapsulate shape information in a manner that is not directly observable and decoding is typically computationally expensive~\cite{Chen2019,Zhang2023_3dshape2vecset}. 
While powerful for generative tasks, the high inference latency and parameter overhead make them less attractive as compact, storable encodings \cite{yu2021plenoctrees,Li-2025-MASH}.

The search for novel 3D representations that 
are able to capture higher geometric details while offering compression levels comparable to those of neural codecs is ongoing in literature. Skeletonization methods such as the medial axis transform \cite{bradshaw2004,Foskey2003,Giesen2009scale_transform,Miklos2010,Sun2016medial_mesh} provide a structural scaffold, typically utilizing unions of spheres to approximate shapes. However, relying on such simple primitives limits their capacity to represent complex geometries.
A more recent work~\cite{Li-2025-MASH}, employs masked spherical functions centered at special anchor points to learn partial distance fields constrained by optimizable viewing cones to the surface. However, adjusting the viewing cones is a challenging task, particularly due to the need to capture details on the border of the corresponding surface patches, a fact that may lead to increased numbers of anchors in cases of complex object shapes. 

In this work, we introduce SHARC, a novel framework that synthesizes complex, genus-agnostic shapes by means of a collection of spherical distance functions 
anchored at selected points in the interior of the object volume. Unlike classical medial axis representations, SHARC utilizes these points as references to encode the distance field to the surface with spherical harmonics. By grounding the representation in internal reference points and explicitly modeling visibility, SHARC enables each SH based distance field representation to capture high-frequency surface details. To identify reference points, we employ a selection algorithm building on \cite{wang2024coverage} that operates on a candidate pool of sampled interior points by short-listing the ones that maximize a score function that encompasses criteria related to surface visibility, as well as centrality and uniformity of short-listed point positions. For each reference point, we capture the surrounding surface geometry via ray-casting and encode the resulting radial distance field using a Fast Spherical Harmonic Transform (FSHT). Our contributions can be summarized as follows:


\begin{enumerate}[label=\alph*)]
    \item \textbf{Execution time}: Both stages of shape encoding and reconstruction are significantly fast in comparison to state of the art methods. 
    \item \textbf{Shape decomposition complexity vs representation accuracy}: The method achieves to strike a balance between the representation complexity (quantified in terms of number of reference points and SH coefficients) and quality of reconstruction that outperforms the state of the art. 

    \item \textbf{Proximity-based surface reconstruction}: Our method learns omnidirectional distance fields for each reference point, allowing the reconstruction to be guided by a proximity-based filtering rule. Instead of blending local patches, SHARC resolves competition between reference points covering the same surface area by prioritizing the nearest point, ensuring a sharper and more accurate reconstruction.
    

\end{enumerate}

\vspace{-0.3 cm}
\section{Related Work}
Below we provide a brief overview of prior works that are relevant to our approach, specifically focusing on skeletal-based, neural, and SH representations.

\subsection{Medial and Skeletal Representations}

Skeletal representations address the problem of reducing shape into a graph structure that spans the original volume. 
Prior research in this domain established that unions of spheres or generalized medial primitives can approximate solids and surfaces with a controllable level of accuracy, tunable via the density and scale of the structural components \cite{bradshaw2004,Foskey2003,Giesen2009scale_transform,Miklos2010,Sun2016medial_mesh}. Subsequent formulations treat skeletal extraction as a global coverage or optimization problem \cite{dou2022coverage,wang2024coverage,Rebain2019_lsmat,Guo2023MSD}, often utilizing greedy selection schemes to prioritize structurally meaningful primitives that satisfy specific geometric constraints \cite{Dou2021}. More recent works augment these ideas with structural and topological guarantees using Restricted Power Diagrams (RPD) - a volumetric partitioning technique that recovers connectivity from medial spheres \cite{Wang2022,Wang2024_mattopo}. Similarly, approaches like Medial Skeletal Diagrams \cite{Guo2023MSD} leverage generalized enveloping primitives to further enhance compactness and reconstruction quality. The community has also begun to exploit medial structures as learnable representations where unsupervised methods infer skeletons directly from point-clouds \cite{Ge2023_point2mm,Lin2021_point2skeleton,Yang2020_P2MAT-Net}. Complementary to skeletal decompositions, recent functional representations model shape via mixtures of local fields anchored at interior reference points, such as Gaussian Process mixtures that learn continuous directional distance functions from sparse data \cite{Sapoutzoglou2026GP}. Our work builds upon the concept of internal geometric representation but moves beyond explicit skeletal structures. Rather than approximating shape through unions of volumetric primitives, we rely on a sparse set of points in the object's interior volume that capture distance fields to the surface with spherical harmonics
, thereby shifting the focus from volumetric approximation to high-fidelity distance field encoding.

\subsection{Neural Implicit representations and Generative Modeling}

Neural implicit representations have been established as a standard approach for compact, high-fidelity 3D geometry. Foundational works \cite{park2019_deepsdf,mescheder2019_occupancy} encode shapes as continuous distance fields parameterized by multilayer perceptrons (MLPs), enabling infinite resolution and efficient compression. Subsequent methods relaxed the reliance on precomputed signed distances by fitting signed distance fields (SDF) directly to raw point clouds or triangle soups, thereby improving robustness to real-world scan data \cite{Atzmon2020,Amos2020}. Other approaches leveraged learnable neural kernels that utilize data-driven priors for improved scalability \cite{williams2021nkf,huang2023nksr}. To handle non-watertight and complex shapes, a range of unsigned and hybrid distance representations have been proposed. These generalize beyond closed surfaces, while maintaining support for accurate surface extraction and geometric queries \cite{Chibane2020,chen2022,Jianglong2022}. Several works extend these implicit encodings with latent shape spaces and multi-scale architectures to support generative modeling and multi-level-of-detail control over continuous 3D geometry \cite{park2019_deepsdf,Chen2019,Zhang2023_3dshape2vecset,zhang2022_2dilg}. Despite the impressive reconstruction fidelity of purely neural approaches, they typically require a lot of time during inference and it is generally difficult to observe how model parameters behave with learning.  
Recently, MASH \cite{Li-2025-MASH} demonstrated that parameterized SH representations can effectively support generative tasks, offering a middle ground between explicit patches and implicit fields. 
Our work, SHARC, takes a similar approach that relies on the more lightweight learning of distance fields from reference points drawn from the interior volume of the shape based on straightforward geometric criteria.   

\subsection{Spherical Harmonic Representations}
Spherical Harmonics have long been used for 3D shape analysis, originally as rotation-invariant shape descriptors \cite{kazhdan2003rotation,papadakis2008efficient}. In the context of full shape representation, global spectral encodings \cite{shen2006sh} are effective for representing star-shaped objects. Recent advancements further refine the reconstruction accuracy of such representations by utilizing Fibonacci grids \cite{Li2024FSH3D} to minimize reconstruction error. However, these single-center methods inherently lack the flexibility to represent shapes with arbitrary topologies with high fidelity. To overcome this, prior methods proposed decomposing complex objects into local, star-shaped patches, allowing efficient SH representation of the resulting over individual sub-parts \cite{mousa2007sh}. More recently, MASH \cite{Li-2025-MASH} advances this local paradigm by utilizing masked, anchored SH fields to support generative tasks. However, MASH relies on a ``multi-view'' surface patching strategy, utilizing generalized view cones and low-order harmonics to tile the surface. This formulation necessitates a large number of anchors and computationally expensive iterative optimization to resolve geometric details. In contrast, SHARC strikes a finer balance between the number of anchors and reconstruction detail by utilizing optimally selected  reference points in the object's interior volume. 

\section{Method}

\begin{figure}
    \centering
    \includegraphics[width=0.99\linewidth]{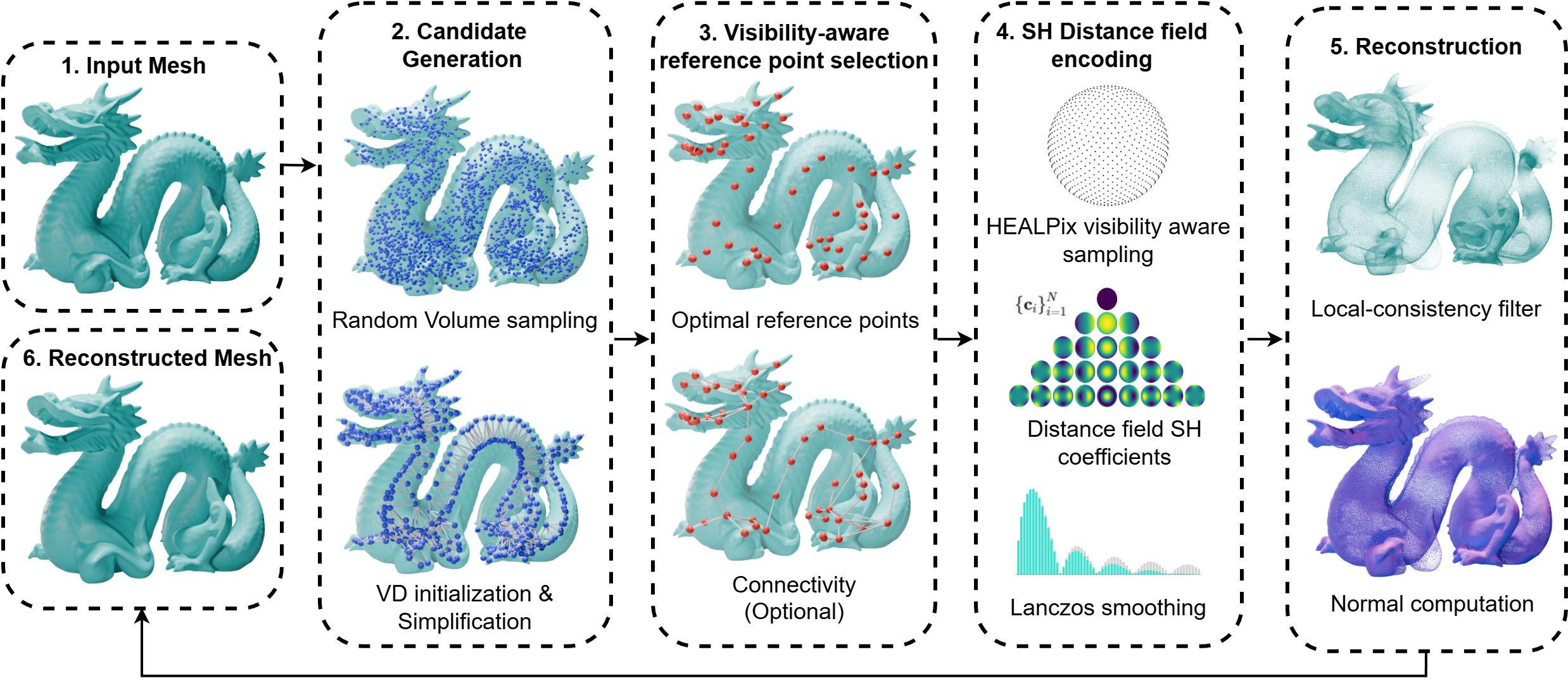}
    \caption{\textbf{Overview of the SHARC pipeline.} From a dense pool of interior candidates, we select a sparse set of reference points that maximize surface visibility. For each point, the local radial distance field is encoded into Spherical Harmonic (SH) coefficients. To reconstruct the shape, we decode these fields and recover the surface points, using a local consistency constraint to filter noisy reconstructed points.}
    \label{fig:method_overview}
    \vspace{-0.5cm}
\end{figure}

Our method aims to produce compact, implicit representations of shapes while striving to maximize the amount of detail carried over to the new representation. To this end, a number of points in the interior volume of the object are employed as reference points from which the surface is learned in the form of distance fields. The overall process comprises three stages: a) \textbf{Candidate generation}, wherein a dense set of candidate reference points is obtained; b) \textbf{Visibility-aware reference point selection}, where we select our reference points as the optimal subset of candidate points that maximizes surface coverage with respect visibility and proximity criteria; and c) \textbf{SH Distance field encoding}, where the distance field to the surface corresponding to each reference point is learned using spherical harmonics. The following sections provide details on the key steps of our method (summarized in Fig. \ref{fig:method_overview}). Detailed algorithms are provided in Suppl. Sec.A.

\subsection{Generation of a dense pool of potential reference points}

A very common strategy to obtain a connected graph of dense points in the interior of the surface is the construction of a Voronoi graph \cite{Amenta2001,li2015qmat,wang2024coverage}. However, this approach is known to be sensitive to boundary noise, where small perturbations of the surface can produce large, unstable features in the skeletal structure \cite{Amenta2001}. Consequently, a raw medial axis is often too complex and detail-heavy to serve as a useful shape abstraction without significant simplification \cite{Miklos2010} (e.g. edge collapse \cite{li2015qmat}). We therefore opt to employ a different strategy that does not require the use of an underlying graph structure.

To obtain a dense pool of potential anchor points located within the object's interior volume,
we initially sample random points inside the surface's enclosing cuboid and thereafter use a ray-casting parity check (also known as Even-Odd Rule \cite{foley1996computer}) to retain the ones that lie in the interior of the object. Particularly the latter step takes advantage of a mesh representation of the original surface and can be regarded as a drop-in replacement for the corresponding step in~\cite{wang2024coverage} employing generalized winding numbers.

In summary, our approach is a more computationally efficient alternative to the standard medial axis transform (MAT) approximation techniques (MAT is the locus of centers of maximal inscribed spheres along with their corresponding radii.~\cite{Amenta2001,li2015qmat,wang2024coverage,dou2022coverage}). 
 
\subsection{Reference point selection} 


Let $\mathcal{M}$ be the two-manifold of the object's surface in $\mathbb{R}^3$. For the final short-listing of reference points we employ a finite set $\mathcal{S} \subset \mathcal{M}$ of points uniformly sampled from the surface $\mathcal{M}$ to aid in the computation of criteria/scores pertaining to the suitability of the anchor points with respect to the reconstruction of the target surface. 

 To ensure sufficient diaspora of the dense anchor-point candidate pool in the interior of $\mathcal{M}$, we employ a centrality and a uniformity score as in~\cite{wang2024coverage}. Specifically, the centrality score favors points in the interior of the object, approximating the Medial Axis and is defined as the inverse of the medial radius,

\begin{equation}
     E_{cen}(\mathbf{c}) = -\frac{1}{\min_{\mathbf{s} \in \mathcal{S}} \|\mathbf{c} - \mathbf{s}\|}
\label{eq:centrality_score}
\end{equation}
where $\mathbf{c}$ is a candidate anchor point and $\|.\|$ denotes the Euclidean norm. The \emph{uniformity} score penalizes concentration by encouraging new anchors to be far from the set of already selected anchors. This is achieved by striving to maximize the minimum distance of the candidate anchor $\mathbf{c}$ from the set of currently selected anchors, $\mathcal{C}_t$ at step $t$, i.e.,
\begin{equation}
    E_{uni}(\mathbf{c}) = \min_{\mathbf{p} \in \mathcal{C}_t} \|\mathbf{c} - \mathbf{p}\|
\label{eq:uniformity_score}
\end{equation}

To promote anchor point placement with respect to optimally covering the target surface, instead of using medial balls or some other geometric primitive, e.g.,~\cite{wang2024coverage}, we express surface coverage based on ray visibility and proximity criteria. This is motivated by the SH representation of a distance field, which requires the surface to be directly visible from the reference point. 
 For a given candidate anchor $\mathbf{c} $ and a surface sample $\mathbf{s} \in \mathcal{S}$ we define the coverage score of $\mathbf{c}$ as,
 \begin{equation}
    E_{cov}(\mathbf{c}) = \frac{1}{\vert \mathcal{S} \vert} \sum_{\mathbf{s}\in \mathcal{S}} \mathrm{Vis}(\mathbf{s}; \mathbf{c})\, \mathrm{Prox}(\mathbf{s}; \mathbf{c}) 
    \label{eq:coverage_score}
\end{equation}
 where, for a point $\mathbf{x}\in\mathcal{M}$ on the manifold, we define $\mathrm{Prox}(\mathbf{x};\mathbf{c})$ and $\mathrm{Vis}(\mathbf{x};\mathbf{c})$ to be the following two criteria as indicator functions:  
\begin{enumerate}[label=\alph*)]
\item \textbf{Proximity},
\begin{equation}
        \mathrm{Prox}(\mathbf{x};\mathbf{c}) = 
        \begin{cases} 
        1\;, &  \|\mathbf{c} - \mathbf{x}\| \leq \tau_{prox} \\
        0\;, & \text{otherwise}
        \end{cases}
 \label{eq:proximity_criterion}
 \end{equation}
 where $\tau_{prox}$ is a configurable distance threshold.

\onehalfspacing
    \item \textbf{Visibility}, 
\begin{equation}
    \mathrm{Vis}(\mathbf{x}; \mathbf{c}) = 
    \begin{cases}
    1\;, & \mathbf{x} = \underset{\mathbf{y}\in\mathcal{M}}{argmin}\,\left\{\| \mathbf{y}-\mathbf{c}\| \,\,\big\vert\,\, \mathbf{y}=\mathbf{c}+\lambda\left(\mathbf{x}-\mathbf{c}\right)\,,\,\, \lambda \in \mathbb{R} \right\}   
    \\
    0\;, & \text{otherwise}
    \end{cases}
    \label{eq:visibility_criterion}
\end{equation}
which determines whether the surface point $\mathbf{x}\in \mathcal{M}$ is visible from $\mathbf{c}$ (i.e., the ray from $\mathbf{c}$ to $\mathbf{x}$ does not cross the surface at any other point).

\end{enumerate}
The coverage score of $\mathbf{c}$ is decided by the conjunction of $\mathrm{Vis}( \mathbf{x}; \mathbf{c})$ and $\mathrm{P}(\mathbf{x}; \mathbf{c})$. This way, abrupt changes in the learned distance field are suppressed, thereby limiting under-fitting by the SH model. 



To obtain the overall score $H(\mathbf{c})$ of $\mathbf{c}$, we combine the scores of Eqs. \eqref{eq:centrality_score}, \eqref{eq:uniformity_score}, \eqref{eq:coverage_score} using a standardization scheme inspired by \cite{wang2024coverage}, 
wherein, following computation of all scores, they are then standardized to yield the final score of each candidate as,

\begin{equation}
    H(\mathbf{c}) = w_{cov} \frac{E_{cov}(\mathbf{c}) - \mu_{cov}}{\sigma_{cov}} + 
                    w_{cen} \frac{E_{cen}(\mathbf{c}) - \mu_{cen}}{\sigma_{cen}} + 
                    w_{uni} \frac{E_{uni}(\mathbf{c}) - \mu_{uni}}{\sigma_{uni}}
                    \label{eq:final_candidate_score}
\end{equation}
where $\mu$ and $\sigma$ denote the mean and standard deviation of the respective scores across the current candidate pool. We typically set weights $w_{cov}=w_{cen}=w_{uni}=1.0$, though these can be tuned to favor sparsity (higher uniformity) or robustness (higher centrality). The candidate with the maximum $H(\mathbf{c})$ is added to $\mathcal{C}_t$, and the set of uncovered points is updated.

\subsection{Learning distance fields with spherical harmonics}
Given the set of selected reference points, $\mathcal{C}$, the surface will be encoded as the aggregate of a number of learned distance fields from each anchor point.  
For each reference point $\mathbf{c} \in \mathcal{C}$, we consider the radial distance function $r(\phi, \theta;\mathbf{c})$ 
that yields the Euclidean distance from the anchor point to the surface along the direction defined by the spherical coordinates $\phi$ and $\theta$ as follows,

\begin{equation}
    r(\phi, \theta;\mathbf{c}) = \underset{\mathbf{y}\in\mathcal{S}}{argmin}\,\left\{\| \mathbf{y}-\mathbf{c}\| \,\,\big\vert\,\, \mathbf{y}=\mathbf{c}+\lambda\,\pmb{\omega}\,,\,\, \lambda \in \mathbb{R} \right\} 
\end{equation}
where $\pmb{\omega}$ is the bearing vector corresponding to the spherical coordinates $(\phi, \theta)$. 

We seek to approximate this function as a truncated Spherical Harmonic expansion up to bandwidth $L$:
\begin{equation}
    r(\theta, \phi) \approx \sum_{\ell=0}^{L} \sum_{m=-\ell}^{\ell} a_{\ell, i}^m Y_{\ell}^m(\theta, \phi)
\end{equation}
where $a_{\ell, i}^m$ are the SH coefficients, and $Y_{\ell}^m$ denotes the spherical harmonic basis function, of degree $\ell$ and order $m$:
\begin{equation}
    Y_{\ell}^m(\theta, \phi) = \sqrt{\frac{2\ell+1}{4\pi}\frac{(\ell-m)!}{(\ell+m)!}} P_\ell^m(\cos\theta) e^{im\phi}
\end{equation}
Here, $P_\ell^m$ are the associated Legendre polynomials. 



To compute $a_{\ell, i}^m$, we sample the sphere using the HEALPix grid~\cite{gorski2005healpix}, which partitions the surface into $N_{pix} = 12 N_{side}^2$ equal-area pixels. Unlike equi-rectangular grids, HEALPix avoids polar distortion and contributes to a more uniform sampling. This structure supports the Fast Spherical Harmonic Transform (FSHT), enabling efficient high-bandwidth learning. We select the appropriate $N_{side}$ to satisfy the sampling lower bounds for bandwidth $L$~\cite{Li2024FSH3D}. We determine the grid resolution in our implementation by targeting a sufficient number of sample points and selecting the nearest valid $N_{side}$.

The grid's equal-area property allows us to approximate the projection integral via discrete quadrature where the term $\frac{4\pi}{N_{pix}}$ represents the constant solid angle $\Omega_{pix}$ of each pixel:
\begin{equation}
    a_{\ell}^m \approx \frac{4\pi}{N_{pix}} \sum_{k=0}^{N_{pix}-1} r_i(\theta_k, \phi_k) \overline{Y_{\ell}^m(\theta_k, \phi_k)}
\end{equation}

\subsection{Reconstruction}

The reconstruction of the shape from the encoded distance fields into a coherent point cloud is a process that includes the following steps: a) Inverse transformation, b) Local generator selection and, c) Normal computation and surface extraction.

\textbf{Inverse transform and Smoothing}. To recover the continuous surface geometry from our representation, we evaluate the stored SH series for each reference point. 
To mitigate occasional minor rippling (see Suppl. Sec. C3) due to discontinuities in the original surface, we employ Lanczos smoothing~\cite{lanczos1956applied}. This technique applies a windowing function to the spectral coefficients, attenuating higher frequencies to ensure smooth convergence without sacrificing global shape fidelity. The filtered coefficients $\hat{a}_{\ell, i}^m$ are computed as:

\begin{equation}
    \label{eq:sh_expansion}
    \hat{a}_{\ell}^m = \sigma_{\ell} \cdot a_{\ell}^m, \quad 
    \text{where} \quad \sigma_{\ell} = \text{sinc}\left(\frac{\pi \ell}{L}\right) = \frac{\sin(\pi \ell / L)}{\pi \ell / L}
\end{equation}

To extract a coherent global surface, we first generate a dense raw point cloud $\mathcal{P}_r$ by decoding the radial distance fields of all reference points. For each reference point $\mathbf{c} \in \mathcal{C}$, we uniformly sample bearing vectors $\pmb{\omega}_j \equiv  (\theta_j,\phi_j)$.  We then evaluate the smoothed radial function to produce surface candidates:

\begin{equation}
    \mathbf{y}(\theta_j, \phi_j; \mathbf{c}) = \mathbf{c} + \tilde{r}(\theta_j, \phi_j; \mathbf{c}) \, \pmb{\omega}_j
\end{equation}

where $\tilde{r}_i$ denotes the radial distance computed using the smoothed coefficients $\hat{a}_{\ell, i}^m$ (Eq. \ref{eq:sh_expansion}). 

\textbf{Local generator selection.} Subsequently, we leverage the association of candidate surface points with their generating reference points in order to filter out artifacts caused by underfitting. Specifically, since spherical harmonics are omnidirectional and encode distance fields constrained only by visibility, the coefficients corresponding to a reference point may generate noisy points in distant regions where angular resolution becomes sparse. To eliminate these artifacts and resolve overlaps, we filter the raw point cloud by retaining a point $\mathbf{y} \in \mathcal{P}_r$ only if its generator is among the $k$-nearest reference points to $\mathbf{y}$.

\textbf{Normal computation and surface
extraction.} To reconstruct the surface, we compute globally consistent normals for the filtered point cloud $\mathcal{P}_{f}$. We first estimate unoriented normals $\mathbf{n}_{raw}$ via Principal Component Analysis (PCA) on a local neighborhood defined by the intersection of $k$-nearest neighbors and a spatial radius $r$.

We resolve the sign ambiguity of PCA without expensive propagation by exploiting our set of internal reference points. Since every point $\mathbf{y}$ originates from a known reference point $\mathbf{c}$, the vector $ \pmb \omega = \mathbf{y} - \mathbf{c}$ provides a reliable outward radial reference. We enforce consistency by aligning the normal with this ray:
\begin{equation}
    \mathbf{n}_y = \text{sgn}(\mathbf{n}_{raw} \cdot \pmb \omega) \, \mathbf{n}_{raw}
\end{equation}
This visibility-based orientation inherently ensures that normals point away from the interior, eliminating the need for minimum spanning tree propagation. Finally, utilizing the oriented point cloud $\mathcal{P}_{f}$ with consistent normals $\mathbf{n}_y$, we generate the final watertight mesh using screened Poisson surface reconstruction~\cite{Kazhdan2013}.

\section{Experiments}
To demonstrate the efficiency and accuracy of our method, we conduct extensive qualitative and quantitative experiments.

\subsection{Experimental Setup}

\textbf{Hardware}. All evaluations are performed on a desktop workstation equipped with an Intel Core i7-13700K CPU, 64GB of RAM, and a NVIDIA GeForce RTX 4090 GPU.

\textbf{Datasets}. We curated a diverse evaluation benchmark comprising models with varying topological complexity from four standard repositories: 16 models from the Princeton Shape Benchmark (PSB)~\cite{shilane2004princeton} representing simple organic shapes (e.g. animals), 6 high-resolution scans from the Stanford 3D Scanning Repository~\cite{stanford3d} featuring dense, high-frequency details, 11 models from Thingi10k~\cite{Thingi10K} exhibiting high genus and topological complexity, and 400 objects from ShapeNet~\cite{chang2015shapenet} (100 per airplane, chair, lamp, and sofa category) to evaluate structured CAD geometry. All input meshes were normalized to fit within the unit sphere prior to processing.

\textbf{Methods}.We evaluate SHARC against three state-of-the-art methods including two skeleton-based approaches, namely the foundational Medial Axis Transform (MAT) and CoverageAxis++ \cite{wang2024coverage} (which optimizes medial spheres for volumetric coverage), as well as MASH \cite{Li-2025-MASH}, which utilizes masked, anchored SH fields to represent geometry. While MSD \cite{Guo2023MSD} was a promising candidate offering high-quality enveloping primitives, we chose to exclude it from this comparison as its optimization time is prohibitively expensive (on average $\sim$30 mins per model~\cite{Guo2023MSD}).

\textbf{Implementation Details}. Unless otherwise noted, we do not enforce a fixed number of reference points for our method, allowing the algorithm to dynamically select a set that maximizes visibility from an initial pool of 10,000 candidates. We set the proximity threshold $\tau_{prox} = 0.2$ across all datasets to balance reconstruction fidelity with storage efficiency (see Suppl., Sec.C1). 
We set the SH bandwidth to $L=64$ (see Suppl. Sec.C2), using $N_{fit} = 10,000$ directions for fitting and $N_{recon} \approx 20,000$ for reconstruction. Throughout the evaluation, we use the term ``primitives'' to collectively refer to the foundational elements of each representation: reference points (Ours), anchors (MASH), and medial balls (Raw MA, CoverageAxis++). For all methods, we configure their parameters for a feasible and fair comparison (see Suppl. Sec.B).

\textbf{Metrics}. To quantitatively evaluate reconstruction accuracy, we employ the L1 Chamfer Distance ($d_{CD}$) and the Hausdorff Distance (HD). Following~\cite{wang2024coverage,Guo2023MSD}, we report the bidirectional HD ($\overleftrightarrow{\epsilon}$), as well as the one-sided components: Ground Truth $\to$ Reconstruction ($\overrightarrow{\epsilon}$) and Reconstruction $\to$ Ground Truth ($\overleftarrow{\epsilon}$). These directional metrics allow us to separately assess completeness and reconstruction precision. Given that all models are normalized to the unit sphere prior to processing, we report these metrics as a percentage of the model's diameter.

\begin{table}[htbp]
\vspace{-0.3cm}
\centering
\caption{Quantitative comparison on the \textbf{PSB}, \textbf{Stanford}, and \textbf{Thingi10k} datasets. We report the primitive count $|C|$, Chamfer Distance ($d_{CD}$), and Hausdorff Distance components $\overrightarrow{\epsilon}$,$\overleftarrow{\epsilon}$, $\overleftrightarrow{\epsilon}$. All metrics are expressed as a percentage of the object's diameter. Best results are highlighted in \textbf{bold}. Detailed per-object results are provided in the Suppl. Sec. D.}
\label{tab:aggregated_results_main}
\resizebox{\textwidth}{!}{%
\begin{tabular}{l|ccccc|ccccc|ccccc}
\toprule
\multirow{2}{*}{\textbf{Method}} & \multicolumn{5}{c|}{\textbf{PSB}} & \multicolumn{5}{c|}{\textbf{Stanford}} & \multicolumn{5}{c}{\textbf{Thingi10k}} \\
\cmidrule(lr){2-6} \cmidrule(lr){7-11} \cmidrule(lr){12-16}
 & $|C|$ & $d_{CD}$ & $\overrightarrow{\epsilon}$ & $\overleftarrow{\epsilon}$ & $\overleftrightarrow{\epsilon}$ & $|C|$ & $d_{CD}$ & $\overrightarrow{\epsilon}$ & $\overleftarrow{\epsilon}$ & $\overleftrightarrow{\epsilon}$ & $|C|$ & $d_{CD}$ & $\overrightarrow{\epsilon}$ & $\overleftarrow{\epsilon}$ & $\overleftrightarrow{\epsilon}$ \\
\midrule
Raw MA & 30k & 0.56 & 2.43 & 2.25 & 2.46 & 83.7k & 0.61 & 3.26 & 2.26 & 3.27 & 53.3k & 2.23 & 7.34 & 6.84 & 7.42 \\
CoverageAxis++ & 310 & 0.62 & 2.86 & 2.20 & 2.87 & 510 & 0.74 & 3.30 & 2.46 & 3.37 & 654$^*$ & 1.59 & 5.53 & 5.26 & 5.83 \\
MASH & 200 & 0.31 & 1.02 & 1.53 & 1.64 & 400 & 0.38 & 1.39 & 1.33 & 1.49 & 400 & 0.52 & 1.77 & 2.74 & 2.99 \\
\textbf{Ours} & \textbf{46} & \textbf{0.28} & \textbf{0.73} & \textbf{0.62} & \textbf{0.74} & \textbf{77} & \textbf{0.33} & \textbf{1.07} & \textbf{0.89} & \textbf{1.08} & \textbf{118} & \textbf{0.39} & \textbf{1.72} & \textbf{1.28} & \textbf{1.78} \\
\bottomrule
\end{tabular}%
}
\vspace{-0.4cm}
\end{table}

\vspace{-0.6cm}
\subsection{Results and Discussion}
\label{sec:results_discussion}

\textbf{Reconstruction Fidelity}.
Table \ref{tab:aggregated_results_main} and Table \ref{tab:shapenet_results} compare our method against state-of-the-art shape representation methods, specifically MASH, CoverageAxis++, and the Raw Medial Axis (Raw MA), across the PSB, Stanford, Thingi10k, and ShapeNet datasets. Our method consistently achieves the lowest Chamfer Distance ($d_{CD}$) across all datasets, while relying on a highly sparse set of primitives ($|C| \approx 45-120$). These are several orders of magnitude 
fewer than Raw MA ($30\text{k}-80\text{k}$) and significantly fewer than MASH ($200-400$) or CoverageAxis++ ($300-650$). We note that CoverageAxis++ failed to produce valid outputs during the connectivity establishment phase for the \textit{Chair} and \textit{City} models in the Thingi10k dataset (denoted by the asterisk in Table \ref{tab:aggregated_results_main}). 
Complex geometries (Thingi10k, Stanford) pose significant challenges for medial-axis-based baselines. Raw MA and CoverageAxis++ exhibit high HD errors ($\overrightarrow{\epsilon}, \overleftarrow{\epsilon}$) due to the inherent nature of the medial primitives, which introduces geometric discontinuities that fail to approximate smooth surfaces accurately. While MASH remains robust on these datasets, it generally yields higher $d_{CD}$ errors than our method despite using up to $4\times$ more primitives (e.g., 0.52 vs 0.39 on Thingi10k). SHARC remains robust across all datasets, maintaining consistently low bidirectional errors ($\overleftrightarrow{\epsilon}$). On CAD models (ShapeNet, Table \ref{tab:shapenet_results}), we observe distinct geometric trends. MASH provides reasonable reconstructions but generally exhibits higher $d_{CD}$ compared to SHARC across all categories. For structured shapes like the \textit{Airplane} and \textit{Chair} categories, SHARC achieves the best Precision ($\overleftarrow{\epsilon}$), significantly outperforming Raw MA. 
Conversely, Raw MA achieves high Completeness ($\overrightarrow{\epsilon}$) in \textit{Airplanes} and \textit{Lamps} due to its dense volumetric coverage whereas SHARC achieves superior Completeness in \textit{Chairs} and \textit{Sofas}. While SHARC can struggle to represent extremely thin, sheet-like surfaces (which require denser reference point coverage, i.e., decreasing $t_{prox}$; see Suppl. Sec. C.1), our method remains superior in overall fidelity ($d_{CD}$) while using less than 0.1\% of the primitive count of Raw MA.

\begin{figure}[t]
    \centering
    \includegraphics[width=0.99\linewidth]{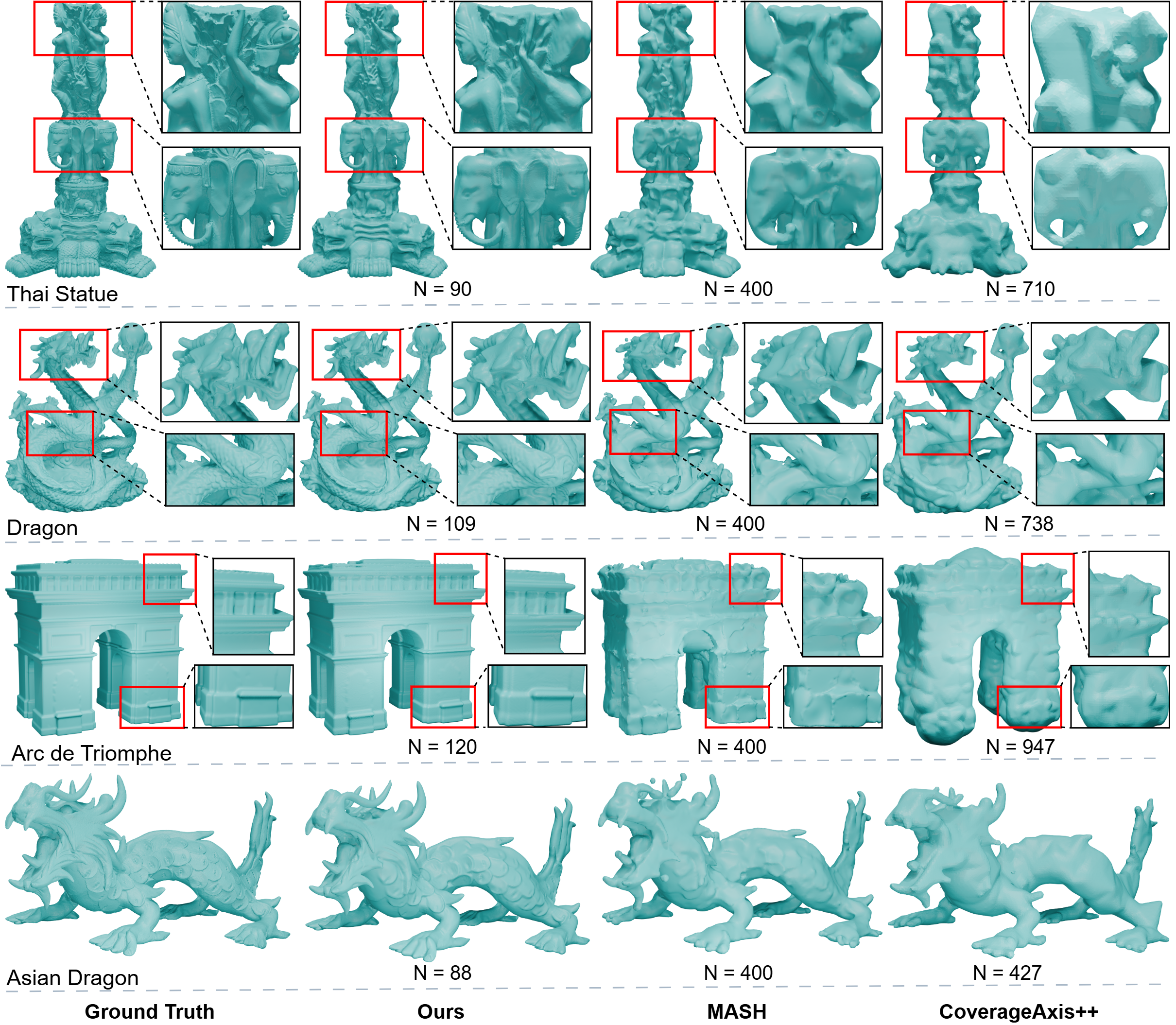}
    \caption{\textbf{Qualitative comparison of reconstruction quality.} We compare SHARC against CoverageAxis++~\cite{wang2024coverage} and MASH~\cite{Li-2025-MASH}, demonstrating our method's ability to preserve fine geometric detail using significantly fewer geometric primitives.}
    \label{fig:qual_results}
    \vspace{-0.5cm}
\end{figure}

\begin{table}[htbp]
\centering
\caption{Quantitative comparison across four ShapeNet Categories (Airplane, Chair, Lamp, Sofa). We evaluate 100 models per category.}
\label{tab:shapenet_results}
\resizebox{\textwidth}{!}{%
\begin{tabular}{l|ccccc|ccccc|ccccc|ccccc}
\toprule
\multirow{2}{*}{\textbf{Category}} & \multicolumn{5}{c|}{\textbf{Raw MA}} & \multicolumn{5}{c|}{\textbf{CoverageAxis++}} & \multicolumn{5}{c|}{\textbf{MASH}} & \multicolumn{5}{c}{\textbf{Ours}} \\
\cmidrule(lr){2-6} \cmidrule(lr){7-11} \cmidrule(lr){12-16} \cmidrule(lr){17-21}
 & $|C|$ & $d_{CD}$ & $\overrightarrow{\epsilon}$ & $\overleftarrow{\epsilon}$ & $\overleftrightarrow{\epsilon}$ & $|C|$ & $d_{CD}$ & $\overrightarrow{\epsilon}$ & $\overleftarrow{\epsilon}$ & $\overleftrightarrow{\epsilon}$ & $|C|$ & $d_{CD}$ & $\overrightarrow{\epsilon}$ & $\overleftarrow{\epsilon}$ & $\overleftrightarrow{\epsilon}$ & $|C|$ & $d_{CD}$ & $\overrightarrow{\epsilon}$ & $\overleftarrow{\epsilon}$ & $\overleftrightarrow{\epsilon}$ \\
\midrule
\textbf{Airplane} & 82k & 0.30 & \textbf{0.90} & 0.58 & \textbf{0.91} & 199 & 0.49 & 2.79 & 1.55 & 2.80 & 200 & 0.27 & 1.21 & 1.47 & 1.63 & \textbf{46} & \textbf{0.23} & 0.92 & \textbf{0.53} & 0.97 \\
\textbf{Chair} & 83k & 0.44 & 1.11 & 0.98 & 1.15 & 200 & 0.72 & 3.49 & 2.34 & 3.52 & 200 & 0.46 & 1.47 & 2.67 & 2.72 & \textbf{99} & \textbf{0.35} & \textbf{0.83} & \textbf{0.89} & \textbf{1.02} \\
\textbf{Lamp} & 71k & 0.32 & 2.20 & \textbf{0.65} & \textbf{2.22} & 167 & 0.54 & 3.79 & 1.63 & 3.87 & 200 & 0.34 & \textbf{1.51} & 2.14 & 2.33 & \textbf{59} & \textbf{0.28} & 2.37 & 0.75 & 2.60 \\
\textbf{Sofa} & 85k & 0.44 & 1.65 & \textbf{0.96} & 1.68 & 200 & 0.79 & 4.41 & 2.61 & 4.41 & 200 & 0.43 & 1.63 & 1.96 & 2.16 & \textbf{116} & \textbf{0.36} & \textbf{0.86} & 1.06 & \textbf{1.19} \\
\bottomrule
\end{tabular}%
}
\vspace{-0.4cm}
\end{table}

\textbf{Qualitative Analysis}. As shown in Fig. \ref{fig:qual_results}, SHARC is able to capture finer details, (e.g., the scales of the \textit{Dragon}) and the facial features of the \textit{Thai Statue} in comparison to MASH and CoverageAxis++. Additional qualitative results can be found in Suppl. Sec. D.

\begin{figure}[htbp]
    \centering
    \includegraphics[width=0.9\linewidth]{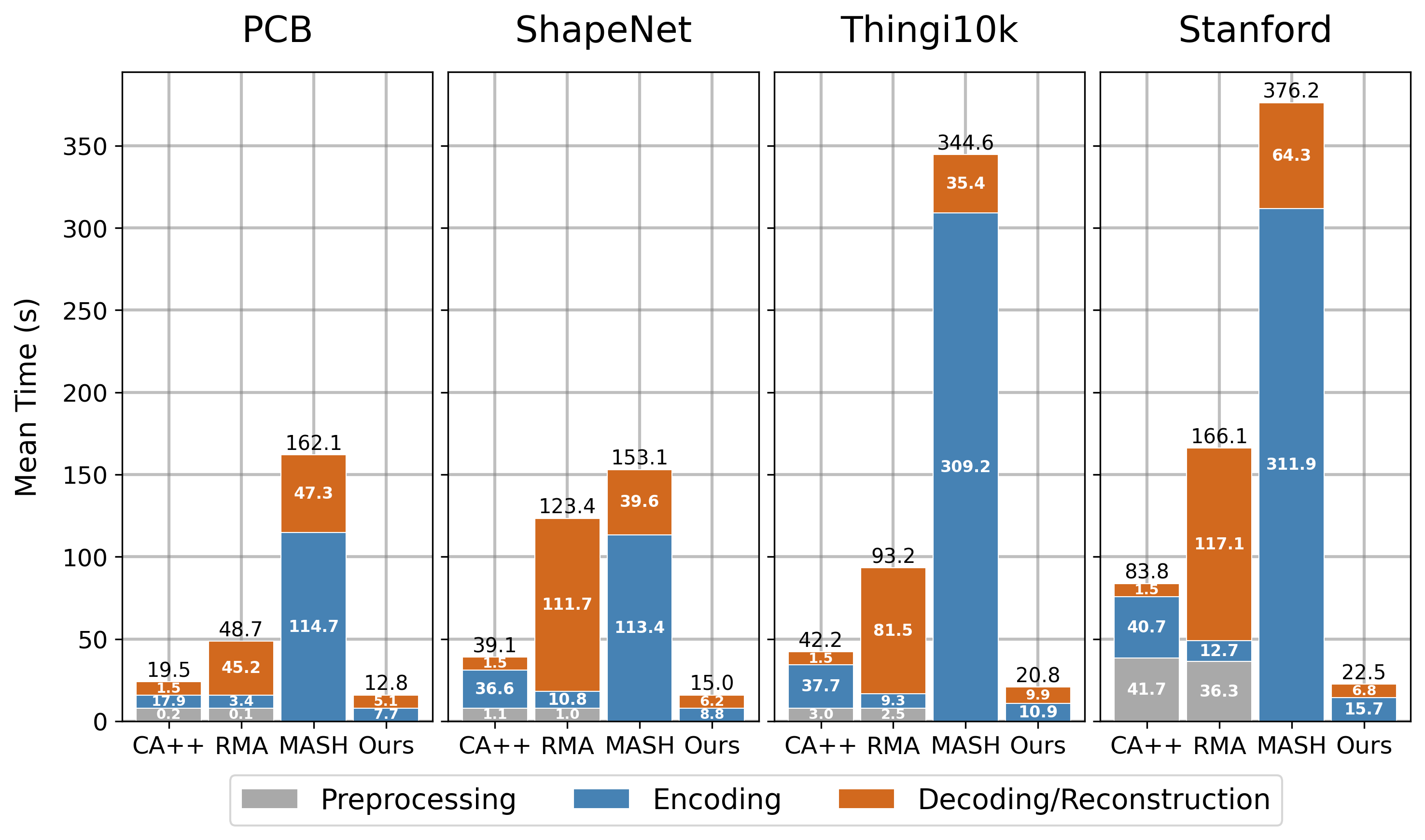}
    \caption{\textbf{Runtime analysis.} Mean execution time (in seconds) for CoverageAxis++ (CA++), Raw Medial Axis (RMA), MASH, and our method across four datasets. The total time is decomposed into Preprocessing (gray), Encoding (blue), and Decoding/Reconstruction (orange).}
    \label{fig:runtime_analysis}
    \vspace{-0.5cm}
\end{figure}

\textbf{Efficiency and Scalability}. Fig. \ref{fig:runtime_analysis} exhibits time complexity of our approach compared to state of the art methods. 
SHARC is able to perform encoding, decoding, and reconstruction efficiently even for very dense and complex geometries (e.g., $\sim$8s to fit coefficients for the \textit{Thai Statue}, a 10M-face mesh). This performance is even more pronounced when comparing, for instance, to the performance of MASH on the Stanford dataset in terms of both accuracy and speed. We provide an ablation study on scalability with increasing primitive count on Suppl. sec C.4.

\begin{table}[htb]
\vspace{-0.3cm}
\centering
\footnotesize
\setlength{\tabcolsep}{2pt}
\caption{\textbf{Storage Efficiency vs. Fidelity.} Mean Compression Ratio ($\rho$, higher is better), Chamfer Distance ($d_{CD}$), and Hausdorff Distances ($\overrightarrow{\epsilon}$, $\overleftarrow{\epsilon}$, $\overleftrightarrow{\epsilon}$) across datasets. Best results are highlighted in \textbf{bold}.}
\label{tab:compression_results}
\resizebox{\textwidth}{!}{%
\begin{tabular}{lccccccccccccccc}
\toprule
\multirow{2}{*}{Method} & \multicolumn{5}{c}{PCB} & \multicolumn{5}{c}{Stanford} & \multicolumn{5}{c}{Thingi10k} \\
\cmidrule(lr){2-6} \cmidrule(lr){7-11} \cmidrule(lr){12-16}
 & $\rho$ & $d_{CD}$ & $\overrightarrow{\epsilon}$ & $\overleftarrow{\epsilon}$ & $\overleftrightarrow{\epsilon}$ & $\rho$ & $d_{CD}$ & $\overrightarrow{\epsilon}$ & $\overleftarrow{\epsilon}$ & $\overleftrightarrow{\epsilon}$ & $\rho$ & $d_{CD}$ & $\overrightarrow{\epsilon}$ & $\overleftarrow{\epsilon}$ & $\overleftrightarrow{\epsilon}$ \\
\midrule
Ours($L$=32) & 1.94 & 0.30 & 0.98 & 0.97 & 1.21 & 153.82 & 0.36 & 1.44 & 1.84 & 1.96 & 9.91 & 0.42 & 2.14 & 1.86 & 2.54 \\
Ours($L$=64) & 0.51 & 0.28 & 0.73 & 0.62 & 0.74 & 40.65 & 0.33 & 1.07 & 0.89 & 1.08 & 2.64 & 0.39 & 1.72 & 1.28 & 1.78 \\
Ours($L$=128) & 0.13 & \textbf{0.27} & \textbf{0.59} & \textbf{0.57} & \textbf{0.61} & 9.85 & \textbf{0.32} & \textbf{0.83} & \textbf{0.78} & \textbf{0.87} & 0.66 & \textbf{0.38} & 1.37 & \textbf{1.17} & \textbf{1.43} \\
Raw Ma & $10^{-3}$ & 0.56 & 2.43 & 2.25 & 2.46 & 0.08 & 0.61 & 3.26 & 2.26 & 3.27 & 0.01 & 2.23 & 7.34 & 6.84 & 7.42 \\
CoverageAxis++ & 0.17 & 0.62 & 2.86 & 2.20 & 2.87 & 32.99 & 0.74 & 3.30 & 2.46 & 3.37 & 2.09 & 1.59 & 5.53 & 5.26 & 5.83 \\
MASH($|C|$=400) & \textbf{20.48} & 0.28 & 0.77 & 0.77 & 0.88 & \textbf{4162.1} & 0.38 & 1.47 & 1.19 & 1.49 & \textbf{276.6} & 0.50 & 1.65 & 2.26 & 2.48 \\
MASH($|C|$=800) & 10.31 & 0.28 & 0.65 & 0.60 & 0.69 & 2096.1 & 0.34 & 0.97 & 0.85 & 0.98 & 139.3 & 0.43 & \textbf{1.24} & 1.78 & 1.85 \\
\bottomrule
\end{tabular}
}
\vspace{-0.4cm}
\end{table}

\textbf{Storage}. As detailed in Table~\ref{tab:compression_results}, SHARC achieves substantial storage reductions compared to dense input meshes. A prime example is the 10M-face \textit{Thai Statue} (Original: 181 MB), which is represented with 1.49 MB ($L=64$)—a $121\times$ reduction—with minimal error ($d_{CD}=0.28\%$). 
In comparison, although MASH ($|C|=800$) achieves excellent compression ratios and reconstruction error comparable to SHARC ($L=64$), 
it however requires more than 10 minutes to reach similar accuracy on the Stanford and Thingi10k datasets. On the other hand, Raw MA is memory-intensive, 
(reaching up to 2 GB for a minimal sphere primitive of 25KB). Finally, CoverageAxis++ reduces the primitive count, but at the cost of lower accuracy.

\section{Conclusion and future work}
In this work, we introduced \textbf{SHARC}, a novel shape representation method that improves and builds on the idea of learning shape as distance fields over skeletal structures. 
Experiments on the PSB, Thingi10k, Stanford, and ShapeNet datasets demonstrate that SHARC outperforms state-of-the-art methods while being significantly faster and more scalable. 
In future work, SHARC can be modified for use as a shape descriptor in recognition tasks. 
Furthermore, the current representation framework can be expanded to include surface texture (e.g., RGB or radiance transfer), creating a unified UV-free representation. Finally, a key research direction would be to make the reference point selection process differentiable in order to include additional geometric criteria in the selection process.

\subsubsection*{Acknowledgement.} This work was partially funded by the European Union’s Horizon Europe programme SOPRANO (GA No 101120990) and PANDORA (GA No 101135775).

\bibliographystyle{splncs04}
\bibliography{references}

\appendix

We provide the source code for SHARC, hosted on GitHub at: \url{https://github.com/pansap99/SHARC}
\section{Algorithmic Details}
\label{app:algorithms}

In this section, we provide the pseudocode for the two core stages of our framework: the spherical harmonic encoding of reference points and the pipeline for surface reconstruction.

\subsection{SH Encoding - SHARC Representation}

Algorithm~\ref{alg:encoding} details the process of converting an input surface $\mathcal{M}$ into the SHARC representation. We first initialize a dense candidate pool and a set of surface samples $\mathcal{S} \subset \mathcal{M}$ (Sec.~3.2). The procedure employs a selection loop to identify the optimal anchor $\mathbf{c}^*$ that maximizes the combined score $H(c)$ (Eq.~6), while respecting the centrality $E_{cen}$, uniformity $E_{uni}$, and visibility-weighted coverage $E_{cov}$ regularization terms. Once the set of reference points $\mathcal{C}$ is determined, we encode the geometry by computing SH coefficients $a_{l}^{m}$ via the Fast Spherical Harmonic Transform (FSHT) on a HEALPix grid.

\begin{algorithm}[h] 
\caption{SHARC Encoding}
\label{alg:encoding}
\SetAlgoLined
\LinesNotNumbered
\small
\KwIn{Surface Samples $\mathcal{S}$, Candidates $\mathcal{P}_{cand}$, Weights $w_{cen}, w_{uni}$, Proximity Threshold $\tau_{prox}$, Sample Directions $N_{fit}$}
\KwOut{SHARC Representation $\mathcal{R} = \{(c, \mathbf{a})\}_{c \in \mathcal{C}}$}

\tcp{Initialization (Sec.~3.1)}
Compute Visibility Matrix $D$ (based on Eq.~5)\;
Compute global Centrality scores $E_{cen}$ (Eq.~1)\;
Initialize Uncovered Set $\mathcal{U} \leftarrow \{1, \dots, |\mathcal{S}|\}$\;
Initialize Selected Anchors $\mathcal{C} \leftarrow \emptyset$\;

\tcp{Greedy Selection Loop (Sec.~3.2)}
\While{$\mathcal{U} \neq \emptyset$ \textbf{and} $|\mathcal{C}| < \text{max\_points}$}{
    \For{each available candidate $c \in \mathcal{P}_{cand}$}{
        Calculate Coverage $E_{cov}(c)$ (Eq. 3) on set $\mathcal{U}$ using matrix $D$\;
        Update Uniformity $E_{uni}(c)$ relative to $\mathcal{C}$ (Eq.~2)\;
        Compute aggregated score $H(c)$ (Eq.~6)\;
    }
    
    Select $c^* \leftarrow \arg\max_{c \in \mathcal{P}_{cand}} H(c)$\;
    
    \tcp{Stop if no new points can be covered}
    \If{$E_{cov}(c^*) = 0$}{\textbf{break}\;}
    
    Add $c^*$ to $\mathcal{C}$ and mark as unavailable\;
    Update Uncovered Set $\mathcal{U}$ by removing points covered by $c^*$\;
}

\tcp{Distance Field Encoding (Sec.~3.3)}
\For{each anchor $c \in \mathcal{C}$}{
    Define local HEALPix grid centered at $c$\;
    Raycast from $c$ to find radial distances $r(\theta, \phi)$\;
    Compute SH coefficients $\mathbf{a}$ via FSHT (Eq.~10)\;
    Store $(c, \mathbf{a})$ in $\mathcal{R}$\;
}
\KwRet{$\mathcal{R}$}
\end{algorithm}

\subsection{Reconstruction}
Algorithm~\ref{alg:reconstruction} describes the surface recovery process. We first apply Lanczos smoothing (Eq.~11) to the coefficients to mitigate ringing. Subsequently, a point cloud $\mathcal{P}_r$ is generated which is then filtered on the basis of generator proximity to ensure geometric consistency (Sec.~3.4). Finally, the normal sign ambiguity is resolved using the visibility vector $\omega = y - c$ (Eq.~13), ensuring all normals $n_y$ point outward before meshing.

\begin{algorithm}[!ht]
\caption{SHARC Surface Reconstruction}
\label{alg:reconstruction}
\SetAlgoLined
\LinesNotNumbered
\small
\KwIn{Representation $\mathcal{R}$, Sample directions $N_{recon}$, Lanczos $\sigma_l$, Neighbors $k$}
\KwOut{Reconstructed Mesh $\mathcal{M}'$}

Initialize raw point cloud $\mathcal{P}_{r} \leftarrow \emptyset$\;

\tcp{Point Generation (Sec.~3.4)}
\ForEach{$(c, \mathbf{a}) \in \mathcal{R}$}{
    Compute smoothed coefficients $\mathbf{\hat{a}}$ using $\sigma_l$ (Eq.~11)\;
    Generate $N_{sample}$ random directions $\{(\theta_j, \phi_j)\}$\;
    \For{each direction $j$}{
        Evaluate radius $\tilde{r}_j$ from SH series $\mathbf{\hat{a}}$ (Eq.~12)\;
        Compute surface point $y = c + \tilde{r}_j \cdot \omega_j$\;
        Add $y$ to $\mathcal{P}_{r}$ tracking its generator $c$\;
    }
}

\tcp{Geometric Consistency Filtering}
\ForEach{$y \in \mathcal{P}_{r}$ with generator $c$}{
    Find $k$-nearest reference points to $\mathcal{N}_k(y)$ in $\mathcal{C}$\;
    \If{$c \in \mathcal{N}_k(y)$}{
        Add $y$ to filtered cloud $\mathcal{P}_{f}$\;
    }
}

\tcp{Normal Computation}
\For{each point $y \in \mathcal{P}_{f}$}{
    Estimate unoriented normal $n_{raw}$ via PCA on local neighborhood\;
    Retrieve parent reference point $c$ of $y$\;
    Define radial visibility vector $\omega = y - c$\;
    Orient normal $n_{y} \leftarrow \text{sgn}(n_{raw} \cdot \omega) n_{raw}$ (Eq.~13)\;
}

\tcp{Meshing}
Generate Mesh $\mathcal{M}' \leftarrow \text{PoissonRecon}(\mathcal{P}_{f}, \{n_{y}\})$\;
\KwRet{$\mathcal{M}'$}
\end{algorithm}

\section{Experimental Setup for other Methods}

To ensure a fair and feasible comparison, we simplify input meshes to 50k faces for CoverageAxis++ and Raw MA, as explicit medial axis computation scales poorly with input complexity. Without this preprocessing, primitive counts become prohibitive (e.g., requiring memory for over 180k spheres for high-resolution inputs), leading to massive memory overhead. For CoverageAxis++, we generate 10,000 interior random samples (similar to our method), bypass raw medial axis simplification (consistent with standard guidelines for random sampling), set dilation to 0.02, and reconstruct the surface via medial ball interpolation after deriving connectivity with QMAT [22].

For MASH, we adapt model capacity to dataset complexity, using 200 anchors for the PSB and ShapeNet datasets and 400 anchors for all others. While we generally follow the authors' evaluation protocol to generate a dense point cloud from the learned representation, we employ Poisson Surface Reconstruction for the final surface extraction. This matches the reconstruction method used in our own pipeline, ensuring that performance differences are attributable to the underlying representation capabilities rather than the meshing algorithm.

\clearpage
\section{Ablation studies}
\label{app:ablations}

To validate the design choices of SHARC, we perform a series of ablation studies evaluating the impact of key hyperparameters on reconstruction quality and representation compactness. 

\subsection{Impact of Proximity Threshold ($\tau_{prox}$)}
A major parameter affecting the density (and total number) of reference points is the interplay between global visibility and the proximity threshold $\tau_{prox}$. As described in Section~3.2, in order for a sampled surface point to be included in the final point cloud, the distance from the corresponding reference point must be below $\tau_{prox}$. Thus $\tau_{prox}$, controls the trade-off between sparsity and reconstruction accuracy (due to proximity to the reference point).

It is important to note that if $\tau_{prox}$ is not set (or set to a sufficiently large value), the selection process is guided almost exclusively by the visibility criterion. In this setting, the algorithm typically selects a smaller optimal subset of reference points required to "see" the entire surface. While this results in a reconstruction with fewer reference points, it will however introduce loss of detail in the reconstruction of distant points from their respective reference points (due to the limited angular resolution of Spherical Harmonics at such distances). 

\begin{figure}
    \centering
    \includegraphics[width=0.99\linewidth]{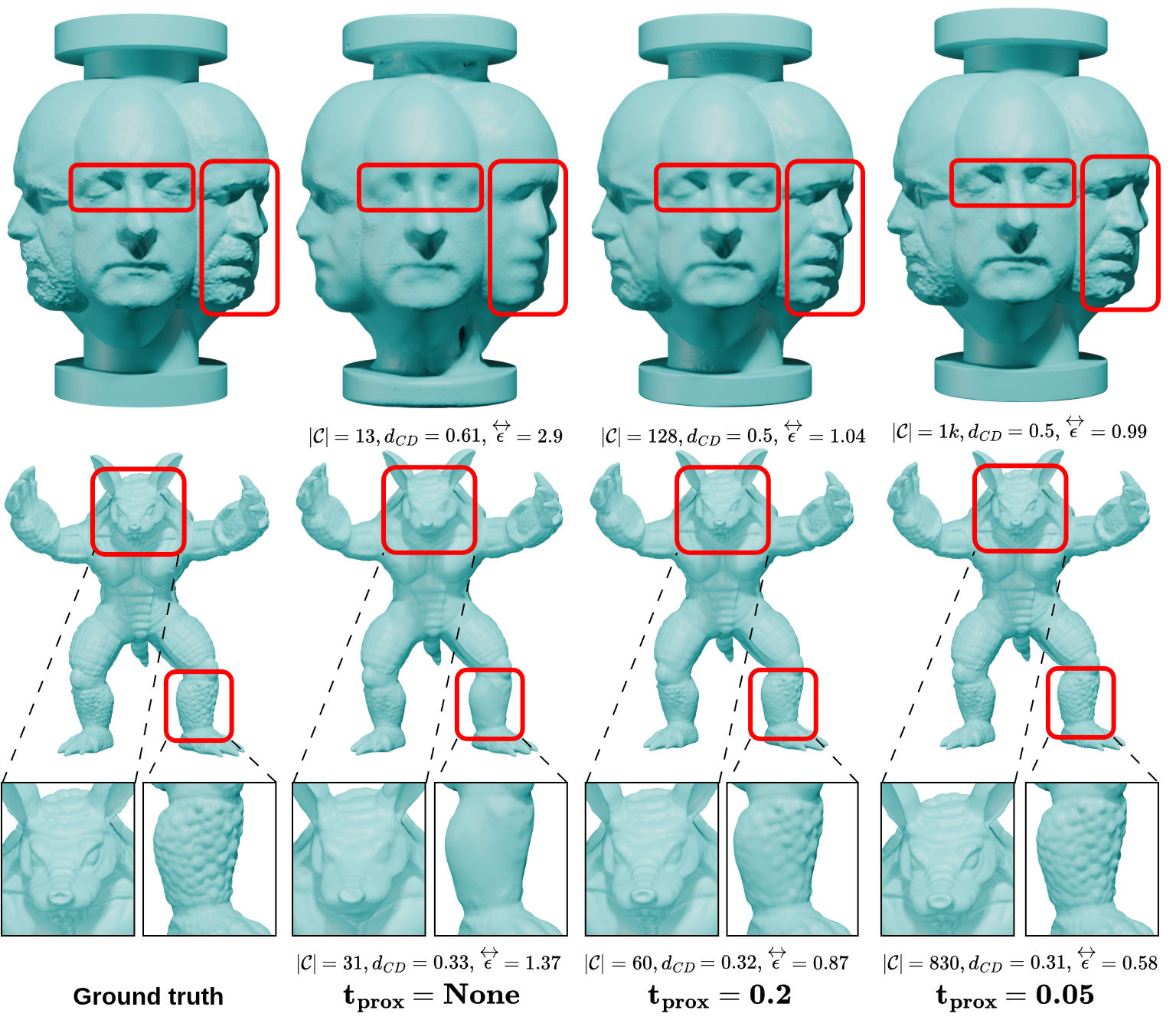}
    \caption{\textbf{Visual Impact of Proximity Threshold ($\tau_{prox}$).} 
    We evaluate reconstruction quality on the \textit{Multi-Face Bust} and \textit{Armadillo} models.
    Without a threshold ($\tau_{prox}=\text{None}$), the selection is driven purely by visibility, resulting in a sparse representation ($|\mathcal{C}|=31$ for Armadillo) that fails to capture fine details like the scales on the leg or facial creases.
    Our default setting ($\tau_{prox}=0.2$) enforces locality, recovering these high-frequency features with a compact set of anchors ($|\mathcal{C}|=60$).
    A strict threshold ($\tau_{prox}=0.05$) results in an excessive number of reference points ($|\mathcal{C}|=830$) without significant visual improvement.}
    \label{fig:ablation_tau}
\end{figure}

In contrast, enforcing a tighter $\tau_{prox}$ forces the algorithm to select more reference points (at the expense of representation parsimony) to satisfy the locality constraint, and therefore improve reconstruction detail. Figure~\ref{fig:ablation_tau} illustrates an example of this behavior. Removing the threshold leads to over-smoothing of fine details (such as scales or facial features) due to insufficient local resolution, while our chosen value of $\tau_{prox}=0.2$ successfully recovers these high-frequency features without the excessive redundancy observed at tighter thresholds.

\subsection{Impact of the SH bandwidth ($L$)}

\begin{figure}
    \centering
    \includegraphics[width=0.99\linewidth]{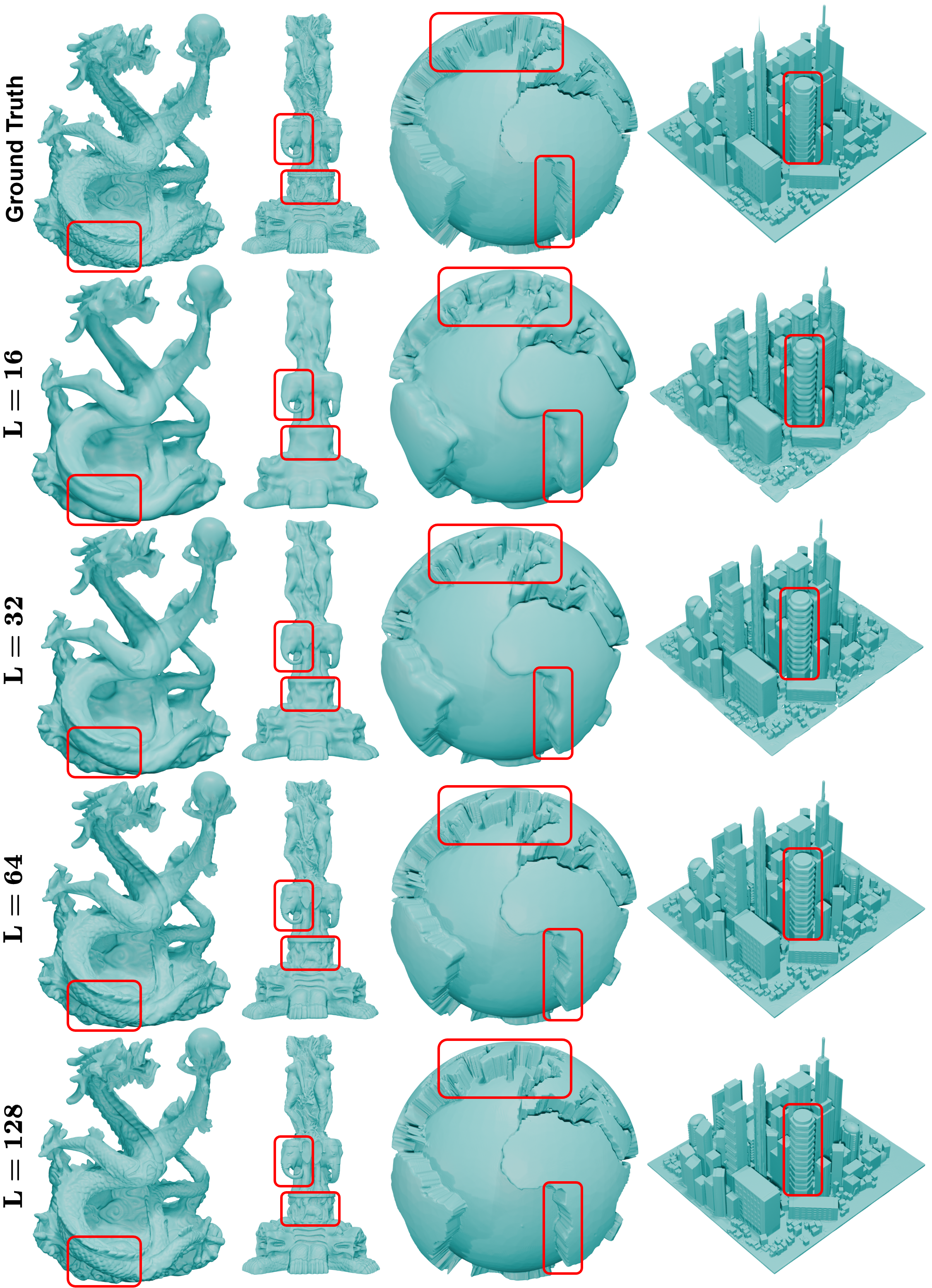}
    \caption{\textbf{Visual Impact of Spherical Harmonic Bandwidth ($L$).} 
    We compare reconstructions of the \textit{Stanford Dragon}, \textit{Thai Statue}, \textit{Earth}, and \textit{City} models at varying bandwidths. 
    Lower bandwidths ($L=16, 32$) result in overly smoothed geometry, failing to capture high-frequency features. 
    Our default setting ($L=64$) successfully recovers fine details such as the dragon's scales and the sharp corners of the skyscrapers. 
    Doubling the bandwidth to $L=128$ offers minor visual improvement while increasing the storage footprint (see Sec.4.2 of main paper).}
    \label{fig:ablation_L}
\end{figure}

We evaluated the reconstruction performance for $L \in \{16, 32, 64, 128\}$. Fig.~\ref{fig:ablation_L} illustrates the impact of this parameter in four different models: \textit{Stanford:Dragon, Thai Statue} and \textit{Thingi10k:Earth,City}.
Setting the bandwidth to lower values ($L=16$ or $L=32$) produces overly smoothed results where high-frequency details are lost, such as the sharp edges of the City buildings or the texture of the Dragon's scales. In contrast, increasing the bandwidth to $L=64$ yields a significant improvement, faithfully preserving these fine structural details and sharp discontinuities. Clearly, further increasing the bandwidth to $L \geq 128$ is plausible, but the cost in the number of SH coefficients becomes significant. Consequently, we utilize $L=64$ for all results presented in this paper as a good trade-off between geometric fidelity and model compactness.

\subsection{Effect of Lanczos Smoothing}

\begin{figure}
    \centering
    \includegraphics[width=0.99\linewidth]{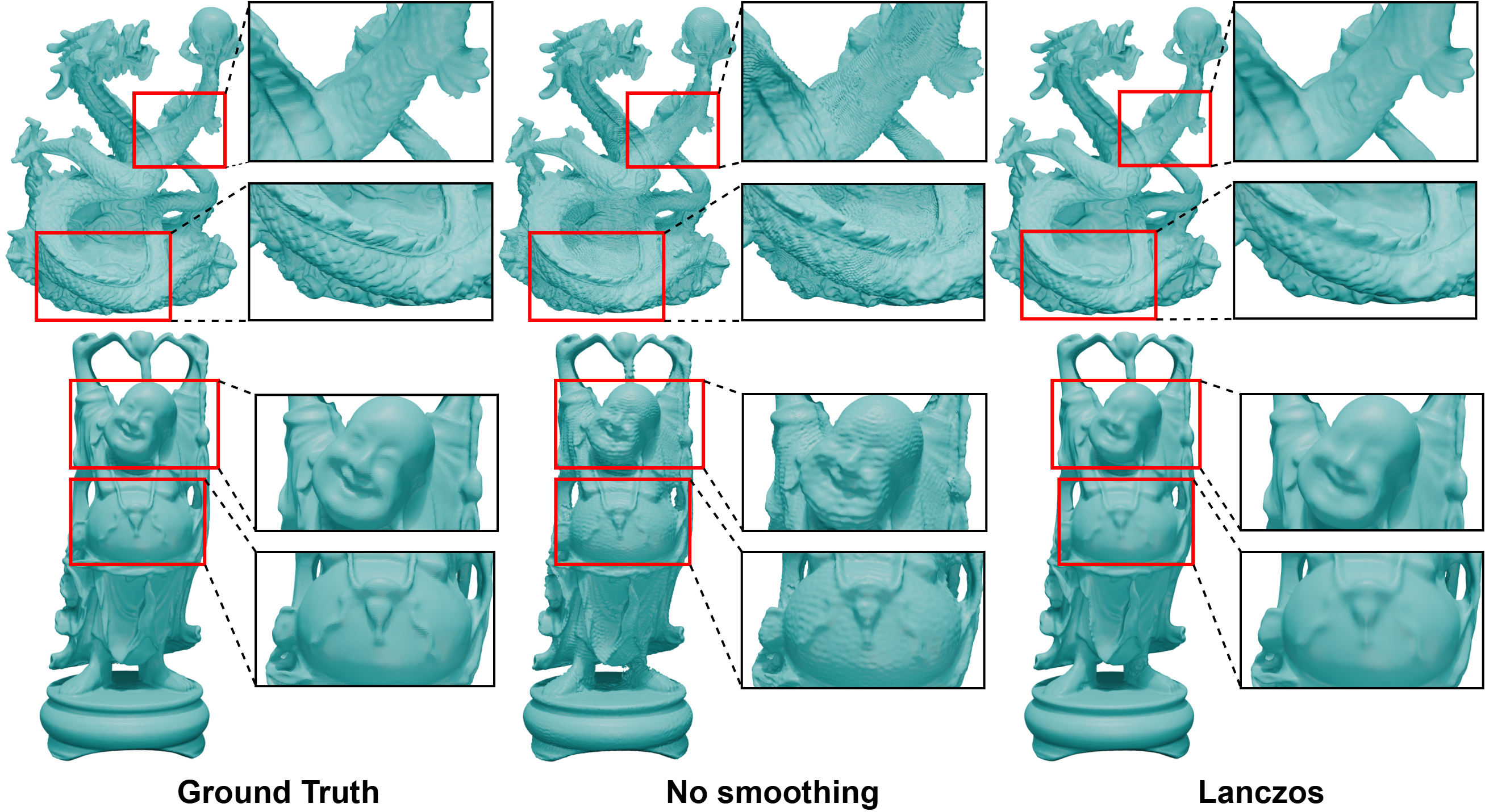}
    \caption{\textbf{Effect of Lanczos Smoothing.} 
    Visual comparison on the \textit{Thingi10k Dragon} model. 
    The \textbf{No smoothing} result exhibits significant ringing artifacts, visible as high-frequency ripples on the neck and body scales. 
    Applying \textbf{Lanczos} smoothing effectively suppresses this noise, restoring a clean surface while preserving the underlying geometric structure. }
    \label{fig:ablation_lanczos}
\end{figure}

Occasionally, primarily due to sampling limitations or abrupt discontinuities in the original surface, underfitting in higher degree harmonics is likely to occur [13,33], typically manifesting as local patch rippling (see middle illustration in Fig. \ref{fig:ablation_lanczos}). 
To mitigate this, we apply Lanczos smoothing (Eq.~11) during the reconstruction phase. This technique applies a $\sigma$-factor window to the spectral coefficients, gradually attenuating the higher frequencies to ensure a smooth convergence of the series.

Figure~\ref{fig:ablation_lanczos} demonstrates the impact of this post-processing step on the \textit{Thingi10k Dragon}. In the "No smoothing" reconstruction, significant ringing artifacts are visible on the neck and body, appearing as unnatural roughness or noise. By applying Lanczos smoothing, these ripples are effectively suppressed, restoring a clean surface that closely aligns with the Ground Truth while preserving the legitimate geometric details of the model.

\subsection{Scalability with Primitive Count ($|C|$)}

\begin{figure}[t]
    \centering
    \includegraphics[width=0.99\linewidth]{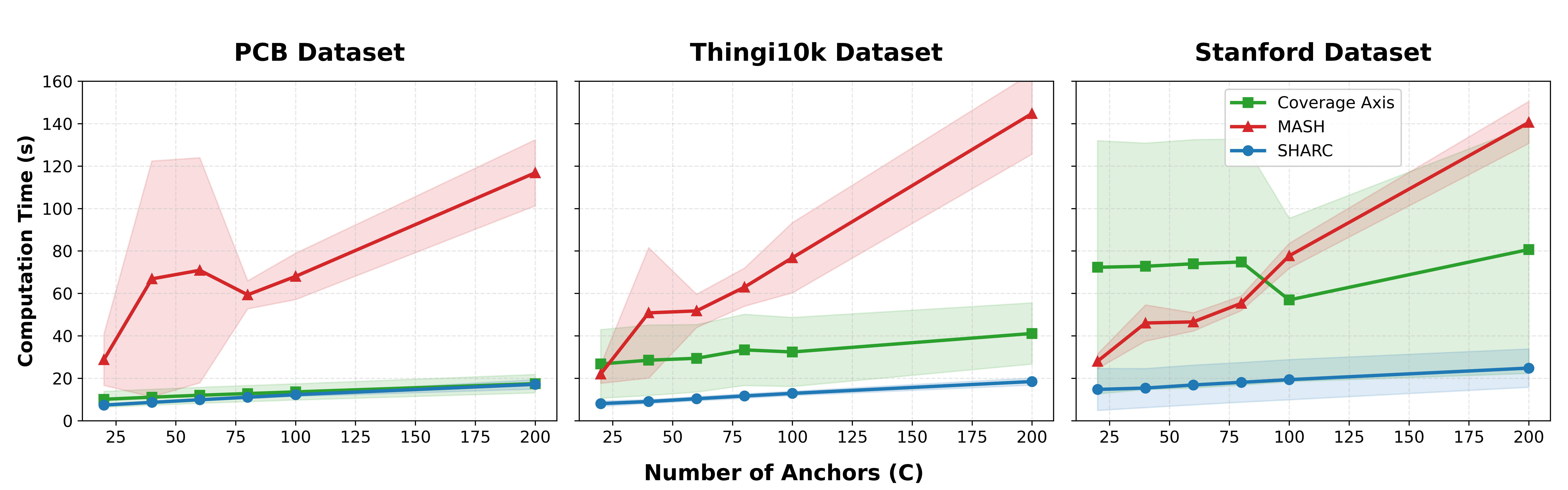}
    \caption{\textbf{Runtime Scalability Analysis.} Comparison of computation time (seconds) vs. number of primitives ($|C|$) across three datasets. \textbf{SHARC} (blue) is consistently fast with minor overhead as $|C|$ increases. \textbf{MASH} (red) exhibits steep linear scaling, while \textbf{Coverage Axis} (green) is fast (for the simplified version of the models; see Sec.B) but offers lower reconstruction fidelity. Shaded areas represent standard deviation.}
    \label{fig:time_comparison}
\end{figure}

\begin{figure}[h]
    \centering    \includegraphics[width=0.99\linewidth]{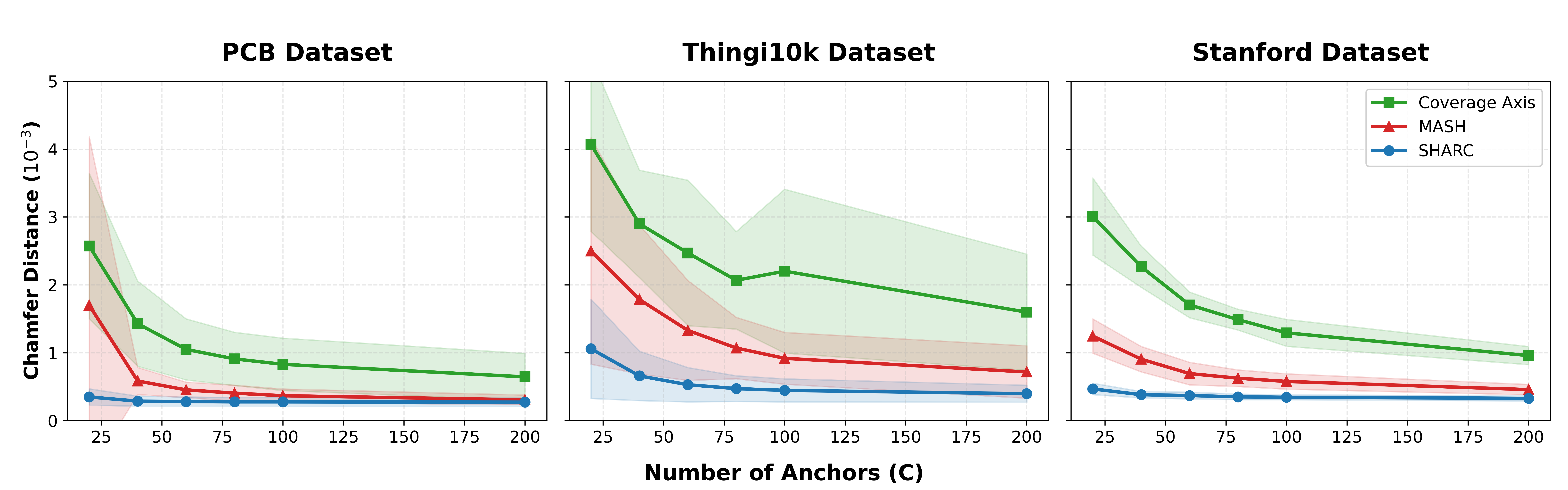}
    \caption{\textbf{Reconstruction Error vs. Primitive Count.} Chamfer Distance ($10^{-3}$) for varying $|C|$. \textbf{SHARC} achieves the lowest error even at the lowest primitive counts, confirming that its speed (shown in Fig. \ref{fig:time_comparison}) does not compromise quality.}
    \label{fig:error_comparison}
\end{figure}

We evaluate the scalability of each method by varying the budget of allocated primitives ($|C|$—comprising medial balls, reference points, or anchors) from 20 to 200. Fig.~\ref{fig:time_comparison} presents the timing results across the PSB, Thingi10k, and Stanford datasets, while Fig. \ref{fig:error_comparison} details the corresponding reconstruction accuracy.

\textbf{Timing Analysis.} As shown in Fig. \ref{fig:time_comparison}, \textbf{SHARC} demonstrates superior efficiency, exhibiting a nearly constant runtime profile with minor overhead as the primitive count increases. Even at $|C|=200$, our method derives the representation in under 20 seconds. In contrast, \textbf{MASH} (red) exhibits a steep linear increase in computation time with significant variance (indicated by the wide shaded regions). \textbf{Coverage Axis} (green) remains competitive but consistently incurs higher computational costs than our method across all datasets.

\textbf{Reconstruction Fidelity.} Simultaneously, the error analysis in Fig. \ref{fig:error_comparison} confirms that SHARC's efficiency does not come at the cost of accuracy. \textbf{SHARC} consistently achieves the lowest Chamfer Distance across all datasets, maintaining high reconstruction fidelity even with a sparse set of reference points ($|C| < 50$). Conversely, \textbf{Coverage Axis} and \textbf{MASH} suffer from significantly higher distortion and instability when the number of primitives is limited, validating the robustness of our method.

\clearpage
\section{Extended Results}

\begin{table*}[htbp]
\centering
\caption{\textbf{Detailed Quantitative Comparison per Model.} We report the primitive count $|C|$, Chamfer Distance ($d_{CD}$), and Hausdorff Distance components ($\overrightarrow{\epsilon}, \overleftarrow{\epsilon}, \overleftrightarrow{\epsilon}$) for individual objects across the \textbf{PSB}, \textbf{Stanford}, and \textbf{Thingi10k} datasets. These values correspond to the dataset means reported in the main paper (Table~1). All metrics are expressed as a percentage of the object's diameter. Best results are highlighted in \textbf{bold}.}
\label{tab:final_results_ordered}
\begin{adjustbox}{max width=\textwidth}
\begin{tabular}{l|ccccc|ccccc|ccccc|ccccc}
\toprule
\multirow{2}{*}{\textbf{Model}} & \multicolumn{5}{c|}{\textbf{Raw MA}} & \multicolumn{5}{c|}{\textbf{CoverageAxis++}} & \multicolumn{5}{c|}{\textbf{MASH}} & \multicolumn{5}{c}{\textbf{Ours}} \\
 & $|C|$ & $d_{CD}$ & $\overrightarrow{\epsilon}$ & $\overleftarrow{\epsilon}$ & $\overleftrightarrow{\epsilon}$ & $|C|$ & $d_{CD}$ & $\overrightarrow{\epsilon}$ & $\overleftarrow{\epsilon}$ & $\overleftrightarrow{\epsilon}$ & $|C|$ & $d_{CD}$ & $\overrightarrow{\epsilon}$ & $\overleftarrow{\epsilon}$ & $\overleftrightarrow{\epsilon}$ & $|C|$ & $d_{CD}$ & $\overrightarrow{\epsilon}$ & $\overleftarrow{\epsilon}$ & $\overleftrightarrow{\epsilon}$ \\
\midrule
\multicolumn{21}{l}{\textit{Dataset: PSB}} \\
\midrule
Ant & 29.7k & 0.33 & 0.65 & 0.72 & 0.72 & 258 & 0.44 & 0.93 & 0.84 & 0.93 & 200 & 0.30 & 0.65 & 0.99 & 0.99 & \textbf{38} & \textbf{0.27} & \textbf{0.50} & \textbf{0.55} & \textbf{0.55} \\
Bird-1 & 25.4k & 0.32 & 0.60 & 0.68 & 0.68 & 325 & 0.39 & 1.04 & 1.02 & 1.04 & 200 & 0.28 & 0.58 & 0.60 & 0.60 & \textbf{28} & \textbf{0.27} & \textbf{0.57} & \textbf{0.53} & \textbf{0.57} \\
Bird-2 & 8.0k & 1.69 & 5.41 & 3.75 & 5.41 & 408 & 1.43 & 5.40 & 3.75 & 5.40 & 200 & 0.29 & 1.82 & 1.42 & 1.82 & \textbf{35} & \textbf{0.26} & \textbf{1.43} & \textbf{1.03} & \textbf{1.43} \\
Bust-1 & 52.9k & 0.51 & 1.80 & 1.82 & 1.82 & 281 & 0.64 & 1.94 & 1.32 & 1.94 & 200 & 0.37 & 1.18 & 2.14 & 2.14 & \textbf{59} & \textbf{0.31} & \textbf{0.54} & \textbf{0.63} & \textbf{0.63} \\
Bust-2 & 17.1k & 1.79 & 13.15 & 13.24 & 13.24 & 564 & 1.52 & 11.13 & 11.15 & 11.15 & 200 & 0.47 & 1.74 & 1.93 & 1.93 & \textbf{99} & \textbf{0.37} & \textbf{1.08} & \textbf{0.79} & \textbf{1.08} \\
Camel & 32.7k & 0.54 & 1.87 & 1.44 & 1.87 & 291 & 0.52 & 3.13 & 1.18 & 3.13 & 200 & 0.30 & 2.05 & 1.29 & 2.05 & \textbf{46} & \textbf{0.26} & \textbf{0.98} & \textbf{0.56} & \textbf{0.98} \\
Cow & 22.2k & 0.41 & 1.79 & 1.84 & 1.84 & 240 & 0.59 & 3.49 & 1.35 & 3.49 & 200 & 0.32 & 1.53 & 1.57 & 1.57 & \textbf{57} & \textbf{0.28} & \textbf{0.66} & \textbf{0.52} & \textbf{0.66} \\
Cup & 44.1k & 0.80 & 6.63 & 6.79 & 6.79 & 620 & 0.72 & 4.83 & 4.87 & 4.87 & 200 & 0.49 & 0.67 & 8.19 & 8.19 & \textbf{39} & \textbf{0.33} & \textbf{0.61} & \textbf{0.63} & \textbf{0.63} \\
Dolphin & 23.8k & 0.28 & 0.56 & 0.64 & 0.64 & 178 & 0.46 & 2.15 & 1.08 & 2.15 & 200 & 0.26 & 0.84 & 0.61 & 0.84 & \textbf{34} & \textbf{0.25} & \textbf{0.50} & \textbf{0.50} & \textbf{0.50} \\
Giraffe & 29.5k & 0.26 & 0.82 & 0.51 & 0.82 & 235 & 0.40 & 2.67 & 1.09 & 2.67 & 200 & 0.22 & 0.77 & 0.68 & 0.77 & \textbf{32} & \textbf{0.20} & \textbf{0.67} & \textbf{0.60} & \textbf{0.67} \\
Glasses & 22.6k & 0.20 & \textbf{0.34} & \textbf{0.35} & \textbf{0.35} & 208 & 0.25 & 0.86 & 0.91 & 0.91 & 200 & \textbf{0.16} & 0.61 & 0.46 & 0.61 & \textbf{28} & 0.17 & 0.81 & 0.84 & 0.84 \\
Hand & 33.1k & 0.30 & 0.84 & 0.78 & 0.84 & 193 & 0.43 & 1.64 & 1.18 & 1.64 & 200 & 0.26 & 0.54 & 0.49 & 0.54 & \textbf{29} & \textbf{0.25} & \textbf{0.49} & \textbf{0.46} & \textbf{0.49} \\
Horse & 26.9k & 0.33 & 0.90 & 0.79 & 0.90 & 232 & 0.52 & 1.89 & 1.27 & 1.89 & 200 & 0.28 & 0.98 & 1.08 & 1.08 & \textbf{40} & \textbf{0.25} & \textbf{0.83} & \textbf{0.50} & \textbf{0.83} \\
Octopus & 25.1k & 0.34 & 0.69 & 0.72 & 0.72 & 225 & 0.43 & 0.98 & 0.94 & 0.98 & 200 & 0.28 & 0.55 & 0.72 & 0.72 & \textbf{35} & \textbf{0.25} & \textbf{0.50} & \textbf{0.54} & \textbf{0.54} \\
Teddy & 37.4k & 0.36 & 1.45 & 0.78 & 1.45 & 183 & 0.63 & 1.87 & 1.86 & 1.87 & 200 & 0.34 & \textbf{0.64} & 0.84 & 0.84 & \textbf{58} & \textbf{0.33} & 0.71 & \textbf{0.62} & \textbf{0.71} \\
Vase & 49.8k & 0.52 & 1.33 & 1.15 & 1.33 & 511 & 0.58 & 1.78 & 1.34 & 1.78 & 200 & 0.40 & 1.24 & 1.47 & 1.47 & \textbf{73} & \textbf{0.37} & \textbf{0.73} & \textbf{0.68} & \textbf{0.73} \\
\midrule
\multicolumn{21}{l}{\textit{Dataset: Stanford}} \\
\midrule
Armadillo & 83.8k & 0.43 & 0.89 & 0.78 & 0.89 & 364 & 0.73 & 2.76 & 2.28 & 2.76 & 400 & 0.36 & 1.37 & 1.59 & 1.59 & \textbf{60} & \textbf{0.32} & \textbf{0.87} & \textbf{0.66} & \textbf{0.87} \\
Bunny & 85.3k & 0.45 & 3.05 & 3.09 & 3.09 & 316 & 0.73 & 2.13 & 2.03 & 2.13 & 400 & 0.34 & 0.63 & 0.84 & 0.84 & \textbf{61} & \textbf{0.33} & \textbf{0.60} & \textbf{0.66} & \textbf{0.66} \\
Dragon & 85.0k & 0.52 & 5.18 & 0.94 & 5.18 & 613 & 0.68 & 5.48 & 1.49 & 5.48 & 400 & 0.43 & 2.06 & 1.71 & 2.06 & \textbf{82} & \textbf{0.38} & \textbf{1.43} & \textbf{0.82} & \textbf{1.43} \\
Happy Buddha & 83.7k & 0.50 & 1.33 & 1.26 & 1.33 & 627 & 0.65 & 2.45 & 1.54 & 2.45 & 400 & 0.37 & 1.39 & 1.12 & 1.39 & \textbf{83} & \textbf{0.33} & \textbf{0.98} & \textbf{0.97} & \textbf{0.98} \\
Asian Dragon & 81.5k & 0.43 & 2.39 & \textbf{0.73} & 2.39 & 427 & 0.66 & 2.93 & 3.21 & 3.21 & 400 & 0.33 & 1.33 & 1.51 & 1.51 & \textbf{88} & \textbf{0.28} & \textbf{1.14} & 0.96 & \textbf{1.14} \\
Thai Statue & 82.7k & 1.30 & 6.70 & 6.73 & 6.73 & 710 & 0.96 & 4.06 & 4.18 & 4.18 & 400 & 0.42 & 1.56 & 1.23 & 1.56 & \textbf{90} & \textbf{0.35} & \textbf{1.38} & \textbf{1.28} & \textbf{1.38} \\
\midrule
\multicolumn{21}{l}{\textit{Dataset: Thingi10k}} \\
\midrule
Octocat & 68.9k & 1.85 & 8.57 & 8.71 & 8.71 & 411 & 1.11 & 3.95 & 4.00 & 4.00 & 400 & 0.29 & 2.08 & 1.10 & 2.08 & \textbf{76} & \textbf{0.26} & \textbf{1.85} & \textbf{0.62} & \textbf{1.85} \\
Arc de Triomphe & 81.9k & 1.56 & 6.94 & 6.95 & 6.95 & 947 & 1.24 & 6.96 & 6.96 & 6.96 & 400 & 1.00 & 2.22 & 8.25 & 8.25 & \textbf{120} & \textbf{0.46} & \textbf{1.60} & \textbf{0.87} & \textbf{1.60} \\
Chair & 19.7k & 5.54 & 12.97 & 13.18 & 13.18 & - & - & - & - & - & 400 & 0.39 & \textbf{1.04} & 2.60 & \textbf{2.60} & \textbf{68} & \textbf{0.36} & 3.23 & \textbf{1.92} & 3.23 \\
Chubby Dragon & 88.9k & 0.51 & 5.09 & 1.23 & 5.09 & 408 & 0.77 & 4.78 & 1.82 & 4.78 & 400 & 0.38 & 1.53 & 1.15 & 1.53 & \textbf{91} & \textbf{0.35} & \textbf{1.22} & \textbf{0.72} & \textbf{1.22} \\
City & 25.8k & 3.76 & 9.17 & 8.35 & 9.17 & - & - & - & - & - & 400 & 0.85 & \textbf{2.78} & 4.49 & 4.49 & \textbf{285} & \textbf{0.54} & 3.64 & \textbf{3.91} & \textbf{3.91} \\
Dragon & 80.2k & 0.61 & 1.91 & 1.07 & 1.91 & 738 & 0.70 & 2.83 & 1.53 & 2.83 & 400 & 0.57 & 1.84 & 3.92 & 3.92 & \textbf{109} & \textbf{0.36} & \textbf{1.09} & \textbf{0.67} & \textbf{1.09} \\
Earth & 51.8k & 2.07 & 8.42 & 8.61 & 8.61 & 670 & 2.36 & 6.14 & 7.41 & 7.41 & 400 & 0.64 & 3.56 & 2.38 & 3.56 & \textbf{238} & \textbf{0.53} & \textbf{2.81} & \textbf{1.98} & \textbf{2.81} \\
Lion Statue & 79.0k & 1.18 & 5.39 & 5.44 & 5.39 & 481 & 1.04 & 5.00 & 6.22 & 6.22 & 400 & 0.31 & 0.94 & 1.20 & 1.20 & \textbf{52} & \textbf{0.29} & \textbf{0.80} & \textbf{0.59} & \textbf{0.80} \\
Multi-Face Bust & 48.1k & 1.97 & 5.37 & 5.51 & 5.51 & 944 & 1.98 & 5.74 & 5.82 & 5.82 & 400 & 0.53 & 1.15 & 2.05 & 2.05 & \textbf{128} & \textbf{0.50} & \textbf{0.91} & \textbf{1.04} & \textbf{1.04} \\
Snowflake & 1.9k & 3.86 & 4.86 & 4.07 & 4.86 & 818 & 3.81 & 4.90 & 4.08 & 4.90 & 400 & 0.47 & 1.58 & 1.38 & 1.58 & \textbf{70} & \textbf{0.36} & \textbf{0.96} & \textbf{0.72} & \textbf{0.96} \\
Spiral Lamp & 39.9k & 1.66 & 12.02 & 12.15 & 12.15 & 473 & 1.32 & 9.48 & 9.51 & 9.51 & 400 & 0.34 & \textbf{0.80} & 1.61 & 1.61 & \textbf{64} & \textbf{0.32} & \textbf{0.80} & \textbf{1.04} & \textbf{1.04} \\
\bottomrule
\end{tabular}
\end{adjustbox}
\end{table*}

\begin{figure}
    \centering
    \includegraphics[width=0.9\linewidth]{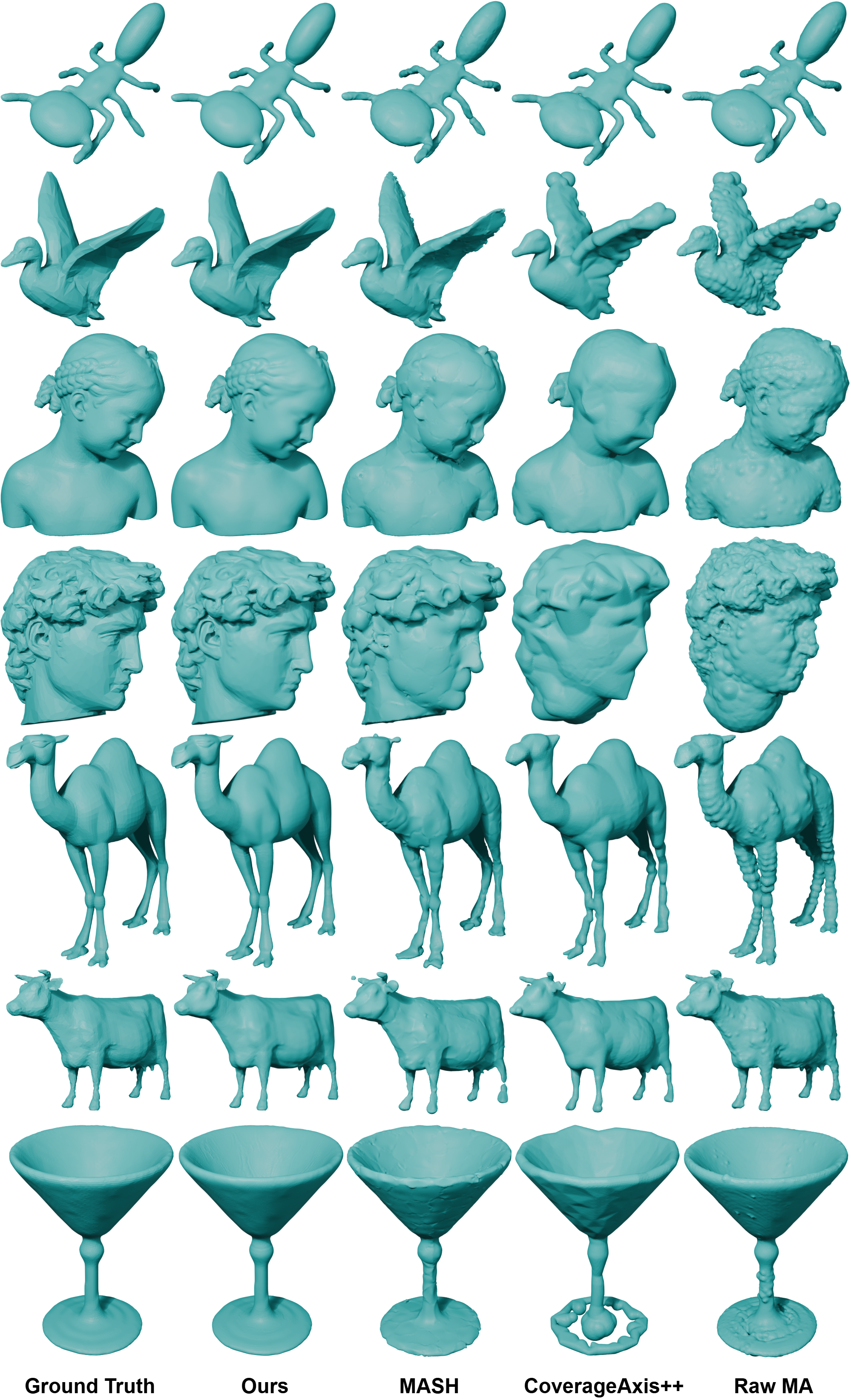}
    \caption{\textbf{Additional Qualitative Comparisons on PSB models.} Visual reconstruction results for the models of Table~\ref{tab:final_results_ordered})}
    \label{fig:placeholder}
\end{figure}

\begin{figure}
    \centering
    \includegraphics[width=\linewidth]{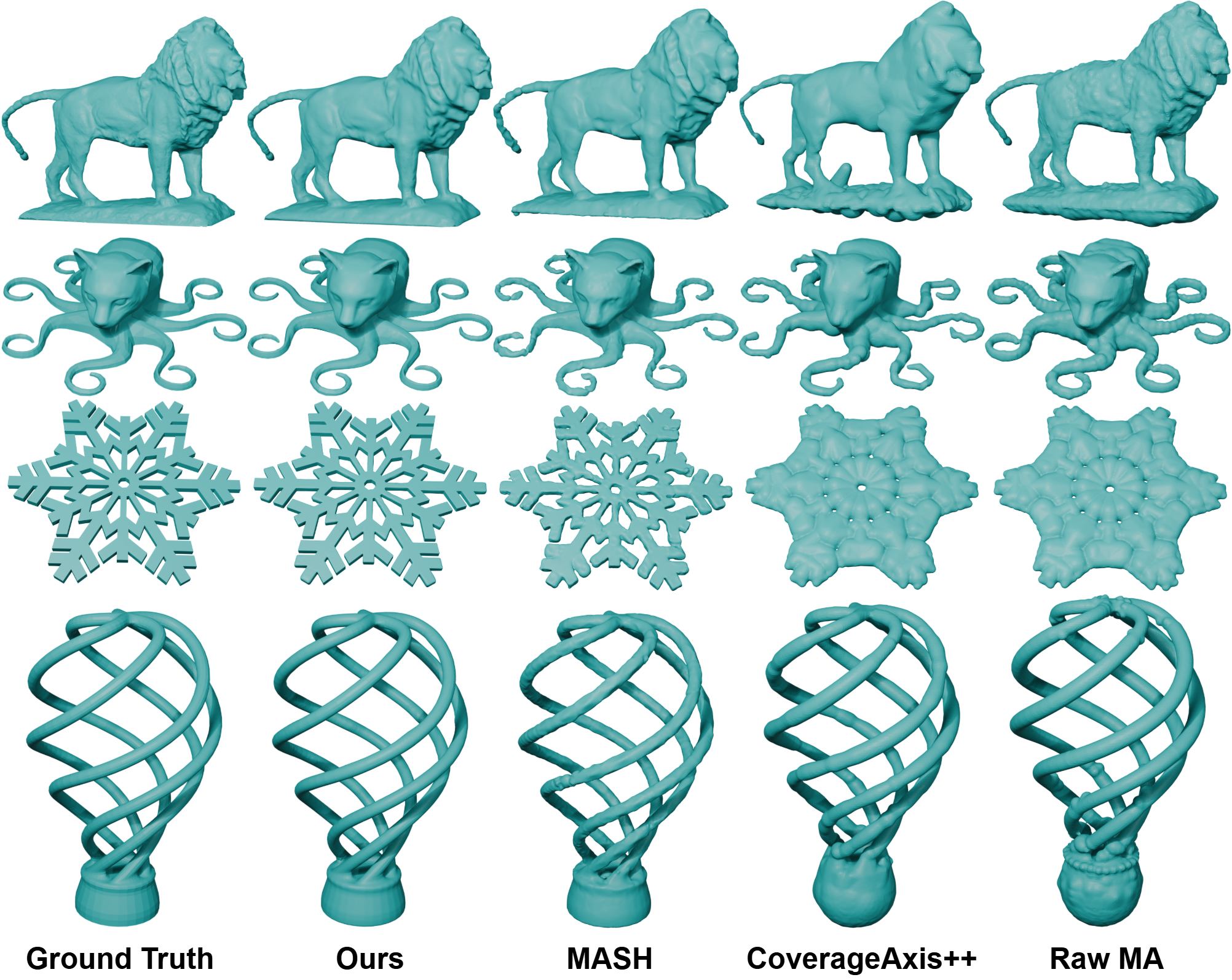}
     \caption{\textbf{Additional Qualitative Comparisons on Thingi10k models.} Visual reconstruction results for the models of Table~\ref{tab:final_results_ordered})}
    \label{fig:placeholder}
\end{figure}

\begin{figure}
    \centering
    \includegraphics[width=\linewidth]{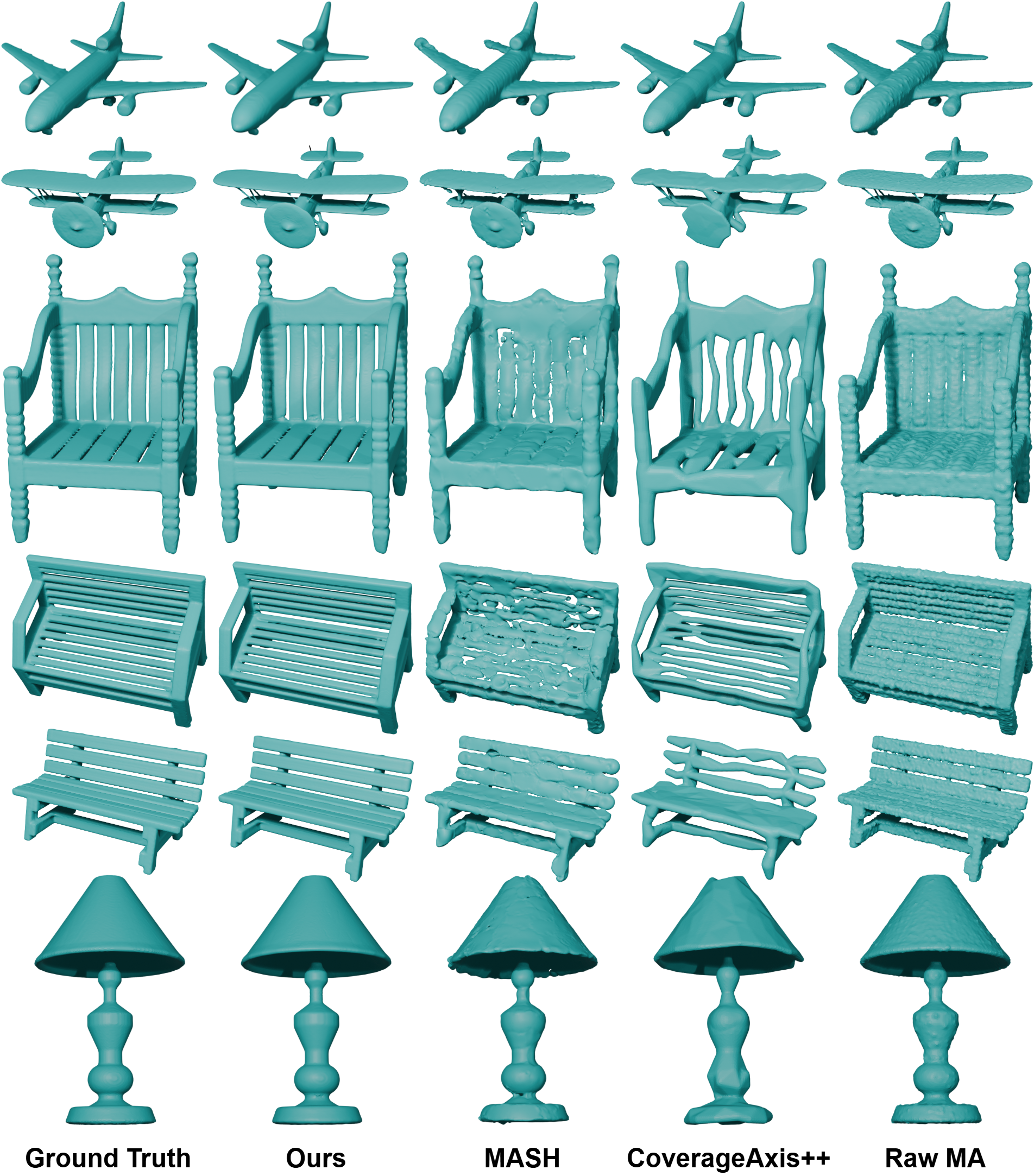}
     \caption{\textbf{Additional Qualitative Comparisons on ShapeNet models.}}
\end{figure}

\end{document}